\documentclass[10pt,twocolumn,letterpaper]{article}

\usepackage{cvpr}              

\usepackage[dvipsnames]{xcolor}

\definecolor{cvprblue}{rgb}{0.21,0.49,0.74}
\usepackage[pagebackref,breaklinks,colorlinks,citecolor=cvprblue]{hyperref}
\usepackage{booktabs}       
\usepackage{amsfonts}       
\usepackage{nicefrac}       
\usepackage{microtype}      
\usepackage{xcolor}         
\usepackage{amsmath,amssymb}
\usepackage{graphicx}
\usepackage{makecell}
\usepackage{amsmath}
\DeclareMathOperator*{\argmax}{arg\,max}

\usepackage{subcaption}
\usepackage{pifont}
\newcommand{\cmark}{\ding{51}}%
\newcommand{\xmark}{\ding{55}}%
\def\paperID{*****} 
\def\confName{3DV\xspace}
\def\confYear{2024\xspace}

\title{NeRF-Feat: 6D Object Pose Estimation using Feature Rendering}

\author{
Shishir Reddy Vutukur$^{1,2}$
 \and Heike Brock$^{2}$ \and Benjamin Busam$^{1}$ \and  Tolga Birdal$^{3}$ \and 
 Andreas Hutter$^{2}$ \and Slobodan Ilic$^{1,2}$ \and \quad \quad
$^{1}$ Technical University of Munich \quad
$^{2}$ Siemens AG \quad
$^{3}$ Imperial College London 
}

\begin{document}
\maketitle
\begin{abstract}
Object Pose Estimation is a crucial component in robotic grasping and augmented reality. Learning based approaches typically require training data from a highly accurate CAD model or labeled training data acquired using a complex setup. We address this by learning to estimate pose from weakly labeled data without a known CAD model. We propose to use a NeRF to learn object shape implicitly which is later used to learn view-invariant features in conjunction with CNN using a contrastive loss. While NeRF helps in learning features that are view-consistent, CNN ensures that the learned features respect symmetry. During inference, CNN is used to predict view-invariant features which can be used to establish correspondences with the implicit 3d model in NeRF. The correspondences are then used to estimate the pose in the reference frame of NeRF. Our approach can also handle symmetric objects unlike other approaches using a similar training setup. Specifically, we learn viewpoint invariant, discriminative features using NeRF which are later used for pose estimation. We evaluated our approach on LM, LM-Occlusion, and T-Less dataset and achieved benchmark accuracy despite using weakly labeled data.     

\end{abstract}    
\section{Introduction}
6D object pose estimation is a crucial component in robotics and augmented reality and is widely deployed in robotic grasping for pick-and-place tasks~\cite{fang2020graspnet,wang2021demograsp,zhai2023monograspnet}. 6D object pose estimation involves estimating rotation and translation from the image reference frame to the object reference frame. 

With the rise in learning-based approaches for solving 6d pose estimation, it has become crucial to develop approaches that work with training data that are easier to obtain without compromising performance. Acquiring real data and obtaining 6D pose annotations requires a highly complex setup and cumbersome effort~\cite{marion2018label,wang2022phocal,jung2022housecat6d}. Rendering synthetic training data using a CAD model yields decent performance, but is still not comparable to the performance obtained using real training data~\cite{denninger2019blenderproc,busam2020like}. 

We propose a pipeline that simplifies the real training data acquisition process and training procedure using weaker labels compared to the strong labels needed and generated by current pipelines. We acquire real training images with relative poses which can be acquired much more easily from the device or using markerboards compared to object pose which requires a complex setup to annotate. We use the relative pose labels to learn the object shape implicitly followed by learning a feature representation of the implicit object shape to enable 6D pose estimation. We employ Neural Radiance Field, NeRF \cite{Nerf}, to learn object shape implicitly from posed real images.     
 
A similar training setup has been employed by NeRF-Pose\cite{Nerfpose}, RLLG\cite{RLLG}. NeRF-Pose is a coordinate regression approach and it cannot handle symmetric objects without knowing symmetric configurations which is the case with the proposed data acquisition pipeline. We propose an approach that takes advantage of 3D representation in NeRF to learn 3D invariant feature space on the surface of the object while respecting symmetries. Specifically, we train NeRF and CNN to learn 3D invariant discriminative feature space by training them together. The learned feature space can be used to establish correspondences between 2D images and learned 3D object representation enabling 6D pose estimation. NeRF encodes the 3D knowledge by design to enable novel view rendering. CNN is invariant to images belonging to symmetric configurations of the object as it only works based on appearance. By training CNN and NeRF together, NeRF transfers 3D knowledge to CNN and CNN forces the features to obey symmetries. This bidirectional learning helps in handling symmetries, unlike NeRF-pose which only transfers 3D knowledge from NeRF to CNN. We employ a contrastive learning approach similar to SurfEmb\cite{surfemb} and Continuous Surface Embeddings \cite{ContSurface} to learn discriminative 3D consistent feature space using InfoNCE loss \cite{infonce}.


To summarize, our contributions are listed as follows:

    
    1. Propose a novel pipeline to distill 3D knowledge to 2D images by rendering features from NeRF.  
    
    2. Providing bidirectional feature learning between CNN and NeRF enables handling symmetric objects.

    3. Simplify the pose estimation training data acquisition and training by working with only relative pose labels and RGB images which are much easier to obtain in practice. 
    
    4. Faster inference on symmetric objects as correspondences are biased to one symmetric configuration when trained with only one object with contrastive learning.

\section{Related Work}

With the rapid prominence of deep learning, learning-based approaches such as \cite{posecnn, densefusion, dpod, DpoDv2, he2020pvn3d, he2021ffb6d, G2lNet, cosypose, surfemb, pix2pose, su2022zebrapose} are proposed which address pose estimation using different input training data modalities and different training data generation pipelines. Dpod\cite{dpod}, DpodV2 \cite{DpoDv2} and Pix2Pose\cite{pix2pose} address the problem of solving pose estimation from RGB images using synthetic data generated from CAD models. BlenderProc \cite{blenderProc} introduced a way to conveniently generate high quality photo-realistic rendering of objects from CAD models using Blender which has improved the accuracy of synthetic data-based approaches. \cite{posecnn, densefusion, he2020pvn3d, he2021ffb6d, G2lNet, gdrn} leverage real training data acquired using complex setup to train the pipeline and have performed better than the approaches trained with synthetic data. 

Recent Methods, Ove6D, OSOP, Latent Fusion, OnePose, OnePose++\cite{ove6d, osop, latentfusion, onepose, oneposeplusplus} propose approaches that can generalize to any object. However, the performance is still not comparable to the approaches using training data from objects under test. To address this issue, Self6D, RLLG, WeLSA, NeRF-Pose, TexPose \cite{self6d,RLLG, welsa,  Nerfpose, texpose} try to use real training data for training by simplifying the acquisition process and by assuming weaker annotation compared to acquiring full 6d real pose annotations which require a complex setup. Self6D pretrains the network using synthetic data from CAD models and then obtains pseudo labels for unlabeled real training data. WeLSA labels a large data of real training images using few labeled real data. RLLG, NeRF-Pose try to learn object pose using relative pose between real training images. Self6D  requires a textured CAD model, RGB-D data and WeLSA requires few labeled RGB-D data to learn to generate labels for real training images. RLLG, NeRF-Pose work with RGB data and weaker labels, but they cannot handle symmetric objects without prior knowledge about the symmetry of the object. We propose an approach to work with weaker labels similar to RLLG and NeRF-Pose and propose an approach that is robust to occlusions and handling symmetric objects without prior symmetry knowledge.   

Neural Radiance Fields, NeRF \cite{Nerf}, proposed a promising way to perform novel view synthesis which has then been adapted to various other tasks like N3F, Distilled feature fields\cite{n3f, distilledfeaturefields}. N3F learns to predict features in addition to colors using a frozen Self-Supervised network, DINO \cite{dino}, to perform scene editing and scene segmentation. In our approach, we also propose to learn features using NeRF, but we can not use a pre-trained self-supervised CNN like DINO to learn features for our NeRF as the learned features are not trained to be invariant to out-of-plane rotations. Besides, Dino is not suited for completely texture-less objects such as T-Less \cite{tless} dataset objects.

Tex-Pose \cite{texpose} uses CAD model to pre-train NeRF and later leverages real data to generate poses for real images by freezing geometry and optimizing texture, and pose for real images.  NeRF-Pose\cite{Nerfpose} and NeRF-Supervision\cite{Nerfsupervision} employ NeRF to estimate correspondences which they use to train another pose estimation network and do not take advantage of robust 3D representation of NeRF for pose estimation. iNeRF \cite{iNerf} acts as a refinement network and needs a good initial estimate to estimate refined pose. NeRF-supervision requires depth maps from Colmap \cite{colmap} in addition to 2D images to train the pipeline. NeRF-Pose uses only 2D images with relative poses, but it cannot handle symmetric objects and employs a unidirectional learning approach from NeRF to CNN. We exploit the 3D knowledge encoded in NeRF to learn 3D invariant features from only 2D images by jointly optimizing NeRF and CNN which in turn enables us to handle symmetric objects.

SurfEmb \cite{surfemb} employs a contrastive feature learning approach that estimates viewpoint invariant features using InfoNCE loss \cite{infonce}. SurfEmb requires a texture-less CAD model and 6D object pose-labeled images for training or a textured CAD model.  We employ a similar contrastive learning approach and additionally make the training procedure simpler by assuming that the relative poses between images are available which are much easier to obtain in practice compared to accurate CAD models and real 6D pose labels. Specifically, we leverage the neural radiance fields, NeRF, to learn object implicit representation, and distill 3D knowledge into 2D CNN while CNN implicitly handles symmetry forcing the features to obey both symmetry and 3D consistency. We address the particularly challenging problem of learning 3D consistent, discriminative features when an accurate CAD model is not present in our pipeline. Learned feature representation is useful for downstream pose estimation tasks by establishing correspondences between 2D images and 3D NeRF representation.


\section{Method}

\begin{figure*}
    \centering
    \includegraphics[width=170mm,scale=1]{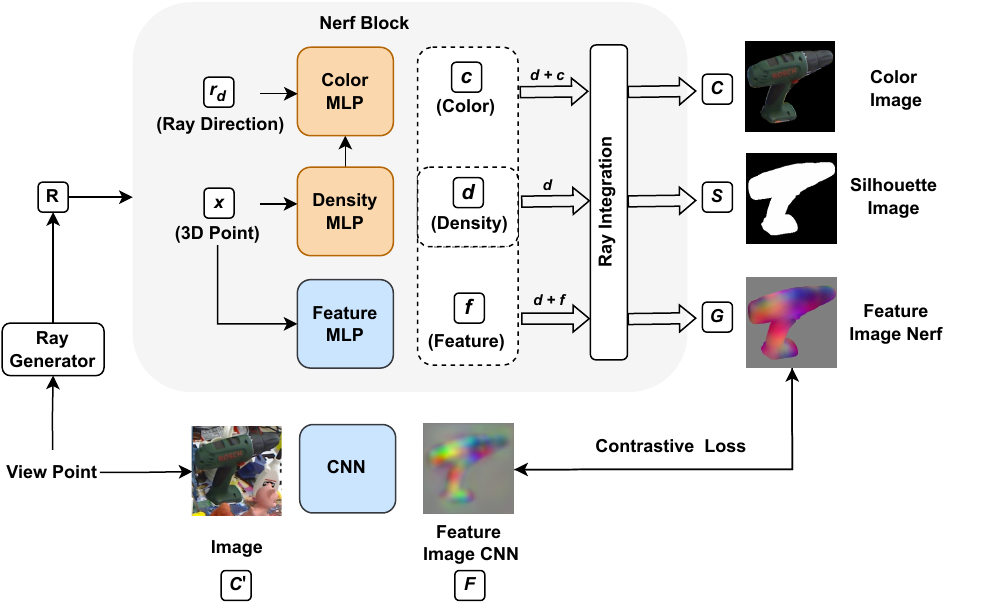}
    \caption{Architecture: Our architecture comprises a Ray Generator,  NeRF Block, and a U-Net CNN.  Ray Generator generates rays from a specific viewpoint which are passed to NeRF Block as discrete 3D points, $x$, along the ray including ray direction, $r_{d}$. NeRF Block comprises three MLPs. Density MLP takes in a 3D coordinate, $x$, and predicts the corresponding density, $d$. Color MLP takes in the intermediate feature from Density MLP and Ray Direction, $r_{d}$ as input to predict color, $c$, at the point. Ray direction is added to color MLP to model view-dependent color changes. Feature MLP takes in a 3D point, $x$, to predict feature vector, $f$. Ray Integration accumulates densities, $d$, and color values, $c$, along a ray to get the final color, $C$. Similarly, density and feature values along a ray are accumulated to generate feature value, $G$, and silhouette value, $S$. Each ray corresponds to a pixel in an image. By generating rays for all the pixels, we can render our final image. Our CNN takes in the input image, $C^{\prime}$ corresponding to the same viewpoint and predicts the feature image $F$. We formulate a contrastive feature loss between feature images from NeRF and CNN.  We train the orange blocks during stage 1 and freeze them during stage 2 when blue blocks are trained. }
    \label{fig:architecture}
\end{figure*}
Our goal is to learn view-invariant per-pixel features from a 2D input image of an object. We employ NeRF to learn an implicit 3D representation of the object using real images with known relative poses without requiring a CAD model. Specifically, we employ a Neural Radiance Field, NeRF, and a U-Net \cite{unet} based CNN to learn 3D consistent feature images. We force CNN and NeRF to learn consistent and discriminative features using contrastive learning. NeRF ensures that the learned features are consistent across viewpoints as it implicitly encodes the 3D information while learning. CNN implicitly handles symmetry as it works only based on appearance and forces the learned features to be similar for visually similar images although from different viewpoints. Moreover, NeRF and CNN compete while learning features as NeRF enforces 3D consistency and CNN enforces similar features to be learned for symmetric configurations.
After the features are learned, we can establish correspondences through features between an image and the 3D model in NeRF which can be used for pose estimation.
We train our pipeline in two stages. We initially train a NeRF network to learn object representation using posed real images. Then, we train CNN and NeRF together to learn features useful for pose estimation. We describe the stages in detail in the following sections:

 \subsection{Stage 1 - NeRF pretraining}
 The training data comprises of real object images and relative poses.  We extract segmentation masks using Segment Any Thing \cite{sam} to extract segmented images. We train the NeRF using segmentation masks and segmented real images to learn object representation. Our training data comprises real images, $C^{\prime}$, segmentation masks, $S^{\prime}$, and relative poses, $T$. 
 NeRF generates rays from the camera center through each pixel of the image to pass through the volume where the object is implicitly reconstructed. NeRF employs a volumetric rendering approach to render images from a viewpoint by shooting rays through the volumetric representation of the object. At each 3D point, $x$, in the volume, NeRF network, $\hat{N}$,  predicts density, $d$, representing the 3D object geometry using Density MLP and also predicts color, $c$ using Color MLP.
 \begin{equation}
    \hat{N}(x) = \left( d, c \right)
 \end{equation}
In addition to rendering color, we render features with our NeRF network, $N$, using another MLP, Feature MLP, to carry out feature learning in the second stage.  So, our network predicts per-point feature, $f$, in addition to density and color as follows:
 \begin{equation}
     N(x) = \left( d, c, f \right)
 \end{equation}
  
 
  \begin{figure}

     \centering
     \begin{subfigure}{.24\linewidth}
     \includegraphics[width=1\textwidth]{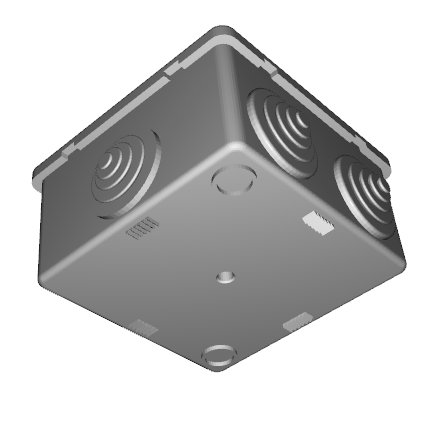}
     \end{subfigure}%
     \begin{subfigure}{.24\linewidth}
     \centering
     \includegraphics[width=1\textwidth]{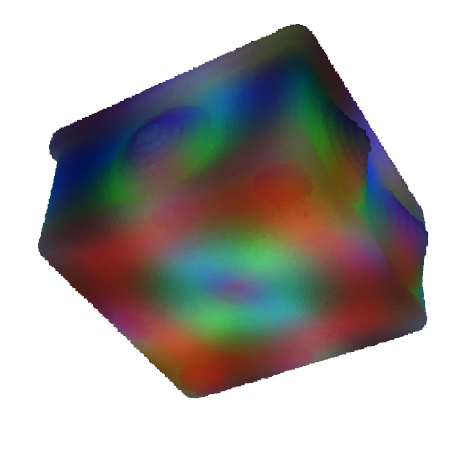}
     \end{subfigure}%
     \begin{subfigure}{.24\linewidth}
     \centering
     \includegraphics[width=0.5\textwidth]{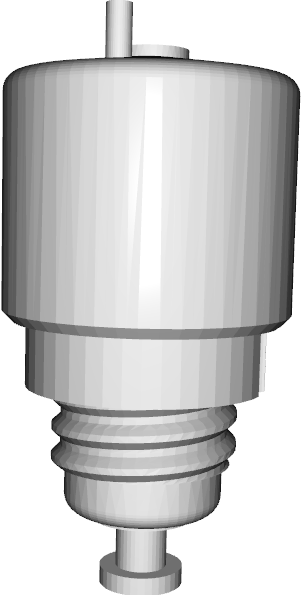}
     \end{subfigure}%
     \begin{subfigure}{.24\linewidth}
     \centering
     \includegraphics[width=0.53\textwidth]{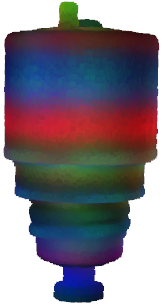}
     \end{subfigure}%

     \caption{Visualization of learned feature representation of symmetric T-Less objects along with meshes }
     \label{fig:featVis}
\end{figure}
 Our architecture is shown in Figure \ref{fig:architecture}. Note that we do not send ray direction as input to Feature MLP as we want to learn view-independent features that resemble the object's intrinsic properties such as albedo and geometry. A ray generated from the camera center through the pixel passes through the volume where the object is being reconstructed. In volumetric rendering, the final color for a specific ray generated for a pixel in the image is obtained by integrating densities and color values along various points along the rays. We query the NeRF network at every point along the ray to obtain per-point density, color and feature. The final color, $C$, feature, $G$, and silhouette, $S$, for a ray with color, $c_{i}$, feature, $f_{i}$, and density, $d_{i}$, at a certain point $i$ along the ray with M points, is computed as follows:
\begin{equation*}
\begin{aligned}
    w_{i}&=\exp(-\sum_{j=1}^{i-1} (-d_{j})) \\
    C&= \sum_{i=0}^{M} w_{i}(1-\exp(-d_{i}))c_{i} \\
    G&= \sum_{i=0}^{M} w_{i}(1-\exp(-d_{i}))f_{i} \\
    S&= 1 -\exp(\sum_{i=0}^{M} d_{i}) 
\end{aligned}
\end{equation*}
 We train our NeRF using the loss formulated between ground truth color image, $C^{\prime}$, segmentation image, $S^{\prime}$, and rendered color, $C$, accumulated using volumetric rendering parameters, density, and colors acquired from NeRF network. We also employ loss between rendered segmentation mask, $S$, and groundtruth segmentation mask, $S^{\prime}$, to formulate silhouette loss. So, the color loss, $L_{c}$, and silhouette loss, $L_{s}$, for $L$ pixels in an image are  formulated as follows:
\begin{equation*}
\begin{aligned}
     L_{c} &= \sum_{i=0}^{L} \|C_{i}-C_{i}^{\prime}  \|_{1} \\
     L_{s} &= \sum_{i=0}^{L} \|S_{i}-S_{i}^{\prime}  \|_{1}
 \end{aligned}    
 \end{equation*}
 
We do not formulate any loss over features as we learn features in the next stage of training. While training NeRF, we optimize both Density MLP and Color MLP to learn geometry through color and silhouette loss. We train the current pipeline to learn robust object representation using the total loss, $L_{1}$ as follows:
\begin{equation}
     L_{1}= L_{c} + L_{s}
 \end{equation}

 \begin{figure*}[h]
    \centering
       \centering
     \begin{subfigure}{.15\linewidth}
     \includegraphics[width=1\textwidth]{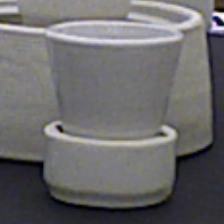}
     \caption{Input Image}
     
     \end{subfigure}%
     \space
     \begin{subfigure}{.15\linewidth}
     \centering
     \includegraphics[width=\textwidth]{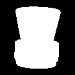}
     \caption{Estimated Mask}
     \end{subfigure}%
     \space
     \begin{subfigure}{.15\linewidth}
     \centering
     \includegraphics[width=\textwidth]{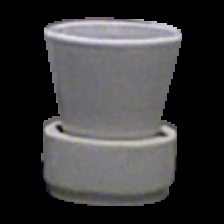}\\
     \caption{Segmented Image}
     \end{subfigure}%
     \space
     \begin{subfigure}{.15\linewidth}
     
     \centering
     \includegraphics[width=\textwidth]{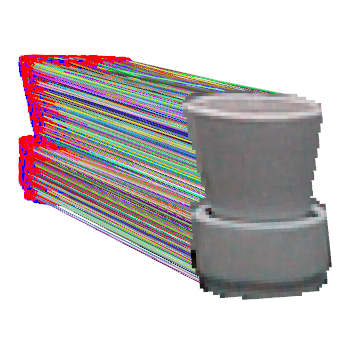}
     \caption{View 1}
     \end{subfigure}%
     \space
     \begin{subfigure}{.15\linewidth}
     \centering
     \includegraphics[width=\textwidth]{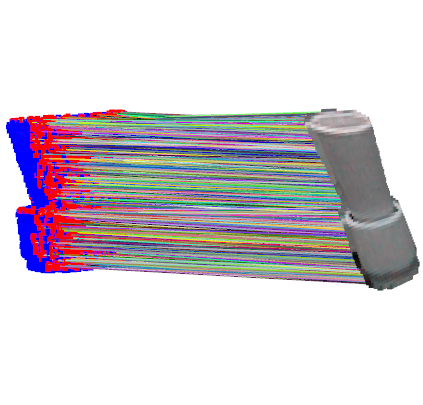}
     \caption{View 2}
     \end{subfigure}%
     \space
    \begin{subfigure}{.15\linewidth}
     \centering
     \includegraphics[width=\textwidth]{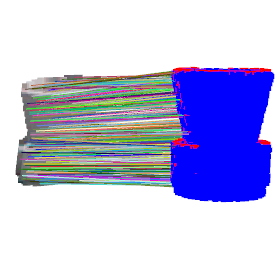}
     \caption{ View 3}
     \end{subfigure}%
    \caption{Correspondence Visualization for continuous symmetric object in T-Less. The Input image, the estimated mask, and the segmented images are shown in the first three images. 2D-3D Correspondences are visualized 
 the next 3 images. Correspondences from 2D segmented region to the 3D point cloud are connected using lines. Blue point cloud is the full point cloud of the object. Red points are the matched correspondences to 2D points. The different views of the correspondence show that the correspondences are biased towards one symmetric configuration. Ideally, for a continuous symmetric object, the correspondences should have been distributed around the object. This bias towards one symmetric configuration helps us in performing inference faster as we can use naive PnP Ransac to estimate the final pose instead of the intensive render and compare inference employed in surfEmb to handle symmetric objects.
}
    \label{fig:CorrVis}
\end{figure*}
\subsection{Stage 2 - Feature Learning}
The training in stage 1 provides an implicit object representation learned in NeRF network. In stage 2, we leverage this learned network and freeze layers which predict the density and color of the 3D point. We freeze the layers so that the geometry of the object is not optimized during feature learning. In this stage, we render features from NeRF for a specific viewpoint based on the relative transformation of the image. We freeze the MLPs trained in the previous stage and optimize the Feature MLP, CNN predicting feature images.


We send an image corresponding to a viewpoint, $k$, as input to U-Net CNN, $U$, to predict a feature image, $F$, at the same resolution as the input image as follows:
\begin{equation}
    F_{k} = U(C^{\prime}_{k})
\end{equation}

Similarly, we render feature image using NeRF from a viewpoint, $k$, using relative transformation, $T_{k}$. For the sake of simplicity, we omit the integration part of the NeRF to get the final feature image from 3d points along the ray. We denote NeRF with integration as $NI$ to generate the feature image, $G_{k}$ from a viewpoint  as follows:
\begin{equation}
    G_{k} = NI(T_{k})
\end{equation}

 We formulate a contrastive loss for learning consistent and discriminative feature representation similar to SurfEmb. To formulate a contrastive loss, we need positive samples and negative samples. Positive samples are pixel-aligned feature pairs between two feature images from NeRF and CNN. To generate negative samples, we shoot rays from all the available viewpoints in training data covering the entire object and render features from different viewpoints. We sample $J$ negative feature samples and denote them as $Z$. We formulate InfoNCE loss which forces positive pairs to be similar and distinctive from negative samples. This helps learn distinctive feature space on the object based on the geometry, and texture of the object. Since we also consider features from CNN, symmetry is automatically handled as viewpoints corresponding to symmetric configurations yield the same feature images. We denote the loss for a specific viewpoint by ignoring $k$ in the notation while formulating the loss. However, the negative samples are independent of the viewpoint as they are sampled from different viewpoints randomly. The loss is formulated for $L$ pixels over CNN feature image, $F$, NeRF Feature Image, $G$, and negative samples, $Z$, as follows:
\begin{equation}
    L_{c}=-\sum_{i=0}^{L} \log \frac{\exp(F^{i} G^{i})}{\sum_{j=1}^{J}\exp(F^{i} Z^{j})}
\end{equation}

Stage 2 is trained with loss, $L_{2}$ for $k$ images as follows:
\begin{equation}
    L_{2}= \sum_{k} L_{c}^{k}
\end{equation}

\subsection{Inference}
After training the pipeline, we established a relation between 2D images and the 3D implicit model through feature space. The learned feature representation is visualized in Figure \ref{fig:featVis}. We leverage this feature space to estimate 6D pose from an RGB image. We shoot rays from different cameras to generate features from different viewpoints and estimate the surface point where the ray hits the surface. We estimate the surface point as the first point along the ray where the density of the point from NeRF network is above a certain threshold. We estimate features at the surface points by passing the 3D points through Feature MLP. In essence, we extract the 3D point cloud with features, $P$, that can be matched with features from CNN feature image. 

For a given image crop, we send the image through CNN to predict the Feature image, $F$. A segmentation mask is applied to the feature image to only focus on the features inside the object region. We estimate the 3D correspondence at the $i^{th}$ pixel in CNN Feature image, $F$,  using point cloud features, $P$, as follows:
\begin{equation}
    \argmax_{j}  \frac{F^{i} P^{j}}{\sum_{j=1}^{J}F^{i} P^{j}}
\end{equation}
The 3D point in the point cloud corresponding to the returned index from the above equation is treated as a correspondence for the 2D pixel. This is evaluated for each pixel in the segmented image to extract many 2D-3D correspondences. We use PnP-Ransac on the estimated 2D-3D matches to extract the final 6D pose. We also perform experiments in case the sensor depth image is available during inference. We render a depth map from NeRF using the pose estimated from the 2D feature image. We adjust the z-axis translation by computing the median difference between a given depth map and rendered depth map. This improves the pose only if the estimated rotation is very good.


\section{Experiments}
We perform experiments on object pose datasets like LineMOD\cite{hinterstoisser2012accv}, LineMOD-Occlusion\cite{occlusion} and T-Less\cite{tless}.  We employ widely used ADD, ADD-S metric to evaluate pose on LineMOD(LM) and LineMOD-Occlusion(LMO) datasets. We employ AR metric from BOP challenge\cite{bop} to evaluate on TLess dataset. We train DpodV2\cite{DpoDv2} segmentation network to extract segmentation masks for inference.
\begin{table*}[!htp]\centering
\caption{LineMOD Results: We present results using RGB, RGB-D data, and CAD model.NP refers to NeRF-Pose with naive PnP ransac,  NP(Mask) refers to NeRF-Pose  with evaluation scheme comparing masks during each ransac iteration}\label{table:linemod}

\scriptsize
\begin{tabular}{lrrrrrrrrr}\toprule
Object &DpodV2\cite{DpoDv2} &GDR-Net\cite{gdrn} &FFB-6D\cite{he2021ffb6d} &LieNet\cite{Lienet} &RLLG \cite{RLLG} &NP\cite{Nerfpose} &NP(Mask) &Ours &Ours \\\midrule
CAD &\cmark &\cmark &\cmark &\xmark &\xmark &\xmark &\xmark &\xmark &\xmark \\\midrule
Data &RGB &RGB &RGB-D &RGB &RGB &RGB &RGB &RGB &RGB-D \\\midrule
ADD &93.5 &93.7 &99.7 &65.2 &82.9 &91.8 &96.6 &\textbf{94.3} & \textbf{99.8}\\
\bottomrule
\end{tabular}

\end{table*}
\subsection{Training Data}
We train our pipeline with images and relative poses. We use Segment Anything \cite{sam} to extract segmentation masks from the images using a bounding box as input to the network along with the image. We assume the object bounding box labels are available if the object is in a cluttered scene as they are much easier to label compared to the segmentation mask. We do not need bounding box labels for objects if the scene only contains the target object as in the T-Less dataset. For inference, we use the 6d pose label of the first frame in our training data to transfer estimated pose to the reference frame of the object defined in the dataset. 
\subsection{NeRF-Pose Comparison}
We compare our approach with NeRF-Pose to show the importance of bidirectional feature based learning instead of employing coordinate regression which is not suited for symmetric objects. NeRF-Pose reports NeRF-Pose(Weak) and NeRF-Pose(Pose). In NeRF-Pose(Weak), they refine the ground truth pose and claim that the accuracy improves the refined poses. Besides, they also propose a mask comparison based intensive ransac where they render and compare segmentation masks with every ransac iteration. They show that they achieve better accuracy using refined poses and intensive ransac. However, these additions are not part of the core approach or core network and can also be applied to our pipeline. To compare the core networks and have a fair evaluation, we compare our approach with NeRF-Pose(pose) since we also use ground truth poses without refinement and also compare with NeRF-Pose with naive PnP ransac which we also employ to estimate pose. We do this to have a fair common ground to evaluate the core network's capability and to show the effectiveness of bidirectional feature learning compared to the correspondence regression based approach. We reimplemented NeRF-Pose on T-Less dataset and also on the LM dataset as they do not provide results with naive PnP ransac. Our approach performs better than NeRF-Pose in LM($2.1\%$), LMO($4.2\%$) and T-Less($4\%$). In a similar setting using ground truth poses and naive PnP ransac, we achieve better accuracy compared to NeRF-Pose.

\subsection{LineMOD}
LineMOD comprises 13 objects of different sizes and textures. We split the dataset similar to previous approaches using $15\%$ data for training and the rest for evaluation. We use ground-truth segmentation during training to have a fair comparison with the NeRF-Pose(NP) pipeline which uses ground-truth segmentation masks for training. However, we perform an ablation study in Table \ref{tab: ablation} with masks generated using Segment Anything to evaluate performance and observed that the accuracy drop is minimal even with generated masks. In Table \ref{table:linemod}, we observe that we achieve closer to benchmark accuracy despite using weaker labels and assuming no CAD model. We observe that our approach performs better than NeRF-pose when we employ naive PnP ransac to evaluate both approaches. However, NeRF-pose(mask) has slightly higher accuracy because of the intensive mask comparison at every ransac iteration for pose estimation which could reduce translation error. The translation error arises mostly along the z-axis component of translation. This is evident from our results using RGB-D data. We render depth using the estimated pose from NeRF network and adjust the translation along the z-axis using the depth map from the sensor by computing the median difference in depth on aligned pixels. This approach will only help if the estimated rotation is very accurate. The results with RGB-D data reflect that our rotation accuracy is much higher and adding depth to the pipeline can help resolve the translation error in some cases. We achieve benchmark accuracy using RGB-D data. Note that the translation adjustment can't be treated as refinement as it is estimated as the median in one shot and not optimized over iterations like ICP \cite{ICP}.


\begin{table*}[!htp]\centering
\caption{LineMOD Occlusion: CAD refers to the approaches using CAD model for training. NP refers to NeRF-Pose and NP(mask) refers to NeRF-Pose with evaluation scheme using mask}\label{table:occlusion}
\scriptsize
\begin{tabular}{lrrrrrrrrr}\toprule
Approach &GDR-Net\cite{gdrn} &ZebraPose\cite{zebrapose} &FFB-6D\cite{he2021ffb6d} &RLLG\cite{RLLG} &NP\cite{Nerfpose} &NP(Mask)\cite{Nerfpose} &Ours &Ours \\\midrule
Modality &RGB &RGB &RGB-D &RGB &RGB &RGB &RGB &RGB-D \\\midrule
CAD &\cmark &\cmark &\cmark &\xmark &\xmark &\xmark &\xmark &\xmark \\\midrule
ADD &62.2 &\textbf{76.9} &\textbf{66.2} &30.3 &44.1 &49.2 &\textbf{48.3} & \textbf{66.9}\\
\bottomrule
\end{tabular}
\end{table*}

\subsection{LineMOD-Occlusion}
LineMOD-Occlusion has a subset of objects in LineMOD with images heavily occluded which tests the robustness of approaches with occlusions. In Table \ref{table:occlusion}, we show that our accuracy is better than NeRF-Pose using PnP ransac without specialized PnP involving projecting mask at every ransac iteration. We also compare with NeRF-pose(pose) reported in their paper instead of NeRF-pose(weak) since we want to compare the core approaches with the same training setup. This shows that our approach performs better in terms of occlusions.
We also achieve higher accuracy after including a depth map in the inference pipeline indicating that the accuracy loss is happening because of translation error. ZebraPose \cite{zebrapose} achieves better accuracy even when compared with approaches using RGB-D data, but its parameterized splitting is not extendable for symmetric objects trivially. Our method slightly lags behind the counterparts using the CAD model which is expected because they have much more accurate synthetic training data generated with natural occlusions in synthetic scenes.  We achieve similar performance to RGB-D based approach FFB-6D\cite{he2021ffb6d} using depth.  

\begin{table*}[h]\centering
\caption{T-Less Results: Evaluation on the T-Less dataset. NP refers to NeRF-Pose approach. We also present AR on continuous symmetric(CS) and Discrete Symmetric(DS) objects on both NeRF-Pose(NP) and our approach. CAD refers to the models using the CAD model for training. Ours-G refers to our model trained without NeRF density network and by querying densities from texture-less CAD model directly to train the pipeline. We report the results of SurfEmb without refinement and Test-Time augmentations for fair comparison.}\label{tab: }
\scriptsize
\begin{tabular}{lrrrrrrrrrrr}\toprule
Appr &DpodV2 &DpodV2 &SurfEmb &NP &NP(CS) &Ours(CS) &NP(DS) &Ours(DS) &Ours-G &Ours \\\midrule
CAD &\cmark &\xmark &\cmark &\xmark &\xmark &\xmark &\xmark &\xmark &\xmark &\xmark \\\midrule
AR &0.65 &0.51 &0.62 &0.54 &0.48 &\textbf{0.52} &0.58 &\textbf{0.63} &0.58 &\textbf{0.58} \\
\bottomrule
\end{tabular}
\label{table:tless}
\end{table*}

\subsection{T-Less}
T-Less has 30 objects comprising 11 continuous symmetric objects, 16 discrete symmetric objects and 3 asymmetric objects. Our approach handles symmetric objects better than NeRF-Pose with a performance gap of $4\%$ as shown in Table \ref{table:tless}. We observe that our approach achieves closer to benchmark accuracy without using CAD models. We performed an ablation experiment(Ours-G) by assuming that a texture-less CAD model is available during training to check the robustness of NeRF representation and performance drop in the absence of the CAD model. We perform this experiment by replacing Density MLP and querying density directly from the CAD model. We observe that there is no drop in accuracy indicating that NeRF is a robust representation and was able to estimate good pose estimates even when some finer details are lost in NeRF compared to the CAD model. Our approach is better suited compared to NeRF-Pose and DpodV2 which are regression-based approaches and need special handling for symmetric objects to achieve better performance. Our approach achieves the best performance in the training setup using real images and pose labels compared to NeRF-Pose and DpodV2. We demonstrate the performance gains using our bidirectional feature learning compared to regression based NeRF-Pose on continuous symmetric and discrete symmetric objects in Table \ref{table:tless}.

\begin{figure}[hb]

     \centering
     \begin{subfigure}{.3\linewidth}
     \includegraphics[width=1\textwidth]{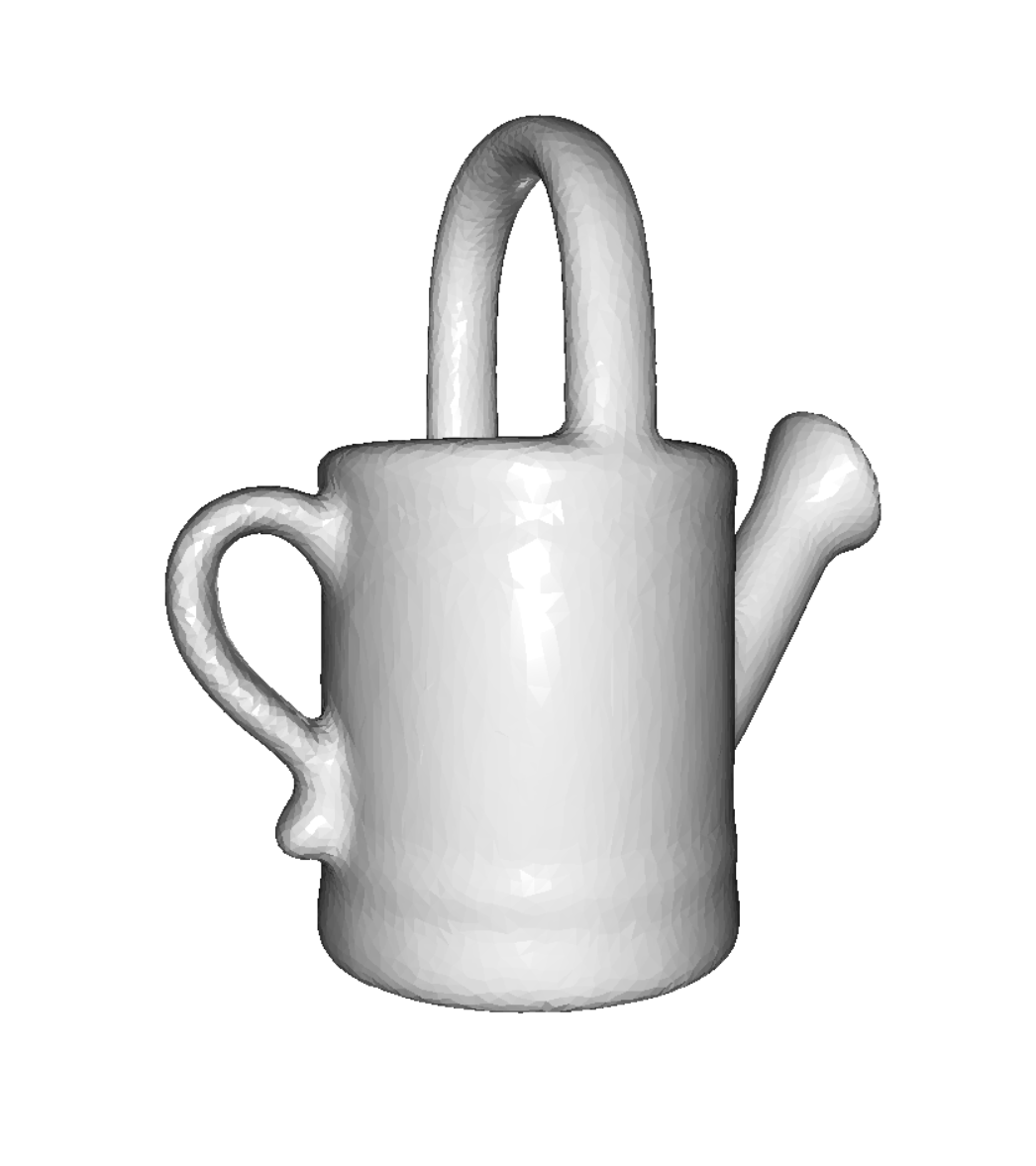}
     \caption{Original mesh}
     \end{subfigure}%
     \space
     \begin{subfigure}{.3\linewidth}
     \centering
     \includegraphics[width=1\textwidth]{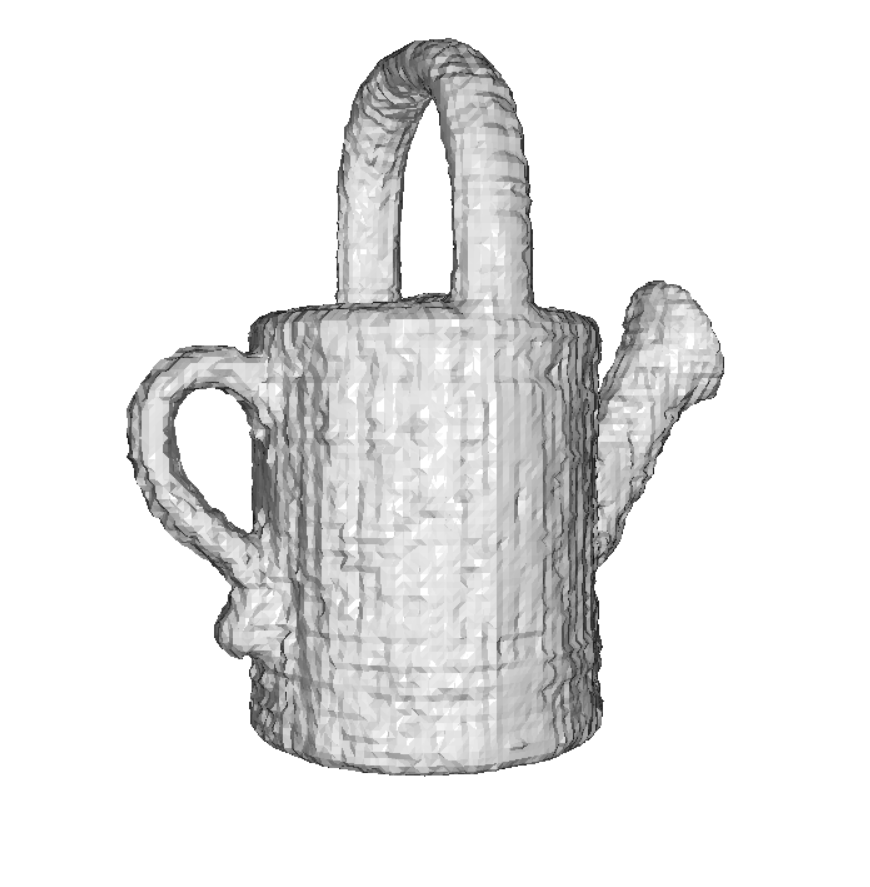}
     \caption{Ours}
     \end{subfigure}%
     \space
     \begin{subfigure}{.3\linewidth}
     \centering
     \includegraphics[width=1\textwidth]{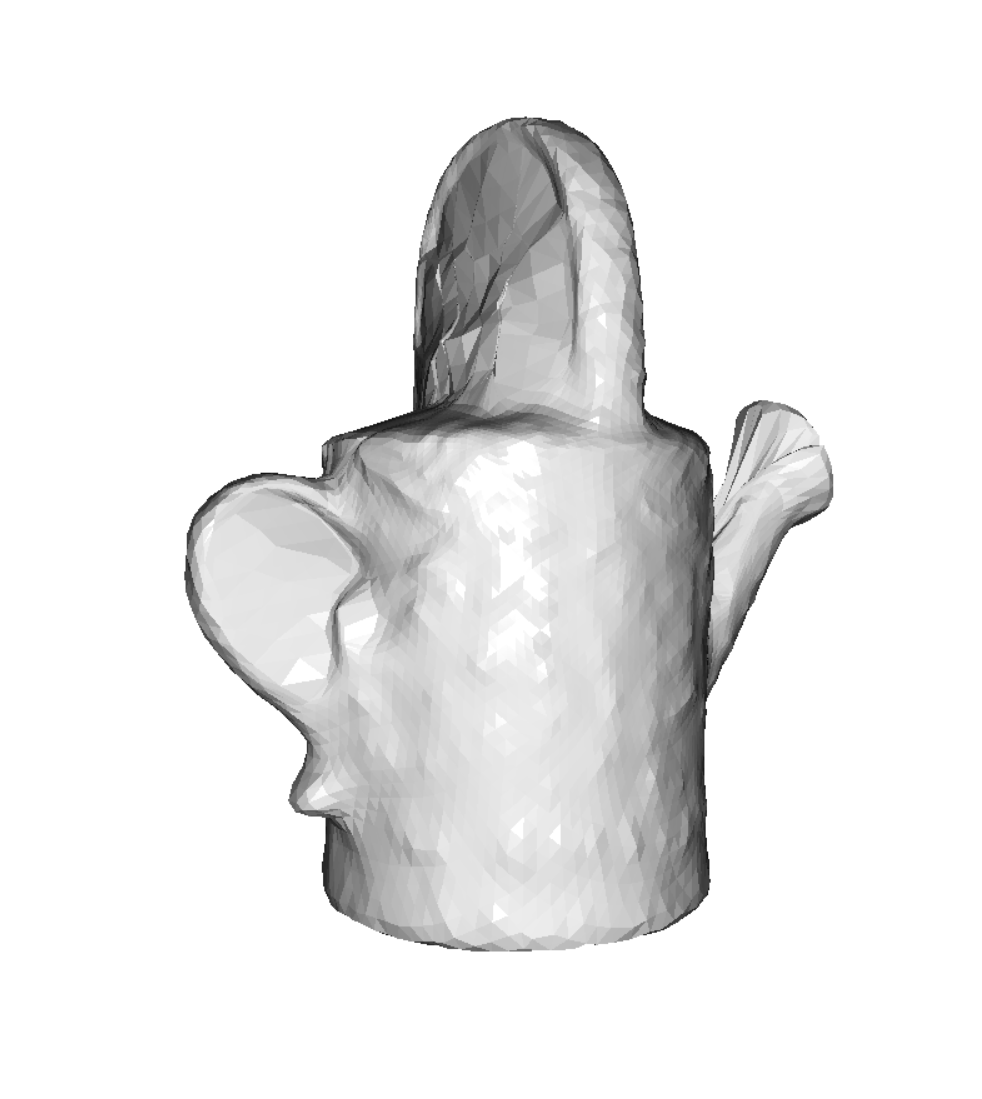}
     \caption{SoftRas}
     \end{subfigure}%

     \caption{Visualization of mesh reconstructions of Can object in LM. The figure shows the original mesh, our reconstruction from NeRF using marching cubes, SoftRas refers to mesh optimized using SoftRas differentiable renderer. SoftRas mesh cannot reconstruct holes as it is optimized from genus zero sphere mesh.}
     \label{fig:meshComparison}
\end{figure}

\subsection{Ablation Studies}
\textbf{Segmentation Mask and Training Views}: We perform ablation studies on three objects from LM. To test the robustness of employing Segment Anything, We consider a non-convex object, Can, and objects with textured patches, Driller and Iron which are difficult to segment and we observed that sometimes the masks were missing some parts in Driller and Iron and sometimes the holes in the masks of Can object were closed. However, we observed that NeRF representation was able to learn a robust object representation despite some wrongly labeled masks. This is clearly reflected in Table \ref{tab: ablation} as there is no drop in accuracy when we use Segment Anything masks compared to ground truth masks. We perform an ablation on the number of training views. We observe that even with just 12 images, we achieve 88\% accuracy which shows that the approach is robust even with a small number of views. We also perform an ablation on one T-Less object to understand the impact of the augmentation pipeline proposed to compensate for the lack of occluded training data. We introduce synthetic occlusions in training data by performing augmentations on object masks and pasting objects on random coco \cite{lin2014microsoft} backgrounds. We observe that our augmentation pipeline boosts the accuracy by 0.32(32\%) in the AR metric.        
\begin{table}[!htp]\centering
\caption{Ablations: We perform 4 experiments with different numbers of views for training and different segmentation masks. SAM refers to segmentation masks generated using Segment Anything and GT refers to using ground truth segmentation masks.}\label{tab: ablation}
\scriptsize
\begin{tabular}{lrrrrr}\toprule
Experiment &Exp1 &Exp2 &Exp3 &Exp4 \\\midrule
Training Views &12 &50 &150 &150 \\\midrule
Segmentation  &GT &GT &SAM &GT \\\midrule
Can &76.8 &93.5 &98.4 &99.0 \\
Driller &91.3 &95.5 &98.3 &98.1 \\
Iron &96.5 &94.2 &99.5 &99.0 \\\midrule
ADD &88.2 &94.4 &98.7 &98.7 \\
\bottomrule
\end{tabular}
\end{table}

\textbf{Robustness of NeRF representation}: We perform an ablation of comparing NeRF with SoftRas \cite{softras} differentiable renderer to emphasize the importance of NeRF in our pipeline. We use a SoftRas-based differentiable renderer from PyTorch3D \cite{pytorch3d} to optimize a mesh based on the pose labels, color images, and segmentation masks. The 3D reconstructions of the mesh are shown in Figure \ref{fig:meshComparison}. Since we do not know the genus of the object beforehand, a differentiable renderer initialized with a zero genus sphere can't reconstruct objects with a genus greater than zero. After reconstructing the mesh, we train the second stage of our pipeline by sampling density from the reconstructed mesh directly instead of using density MLP from NeRF. We evaluate the pose estimation accuracy on LM and LMO in Table \ref{tab: ablation2}. The performance is similar on LM, but we observe that the NeRF based approach is better on the LMO dataset compared to the differentiable renderer based approach. Also, the mesh reconstructions show that NeRF clearly learns better geometry when the genus of the object is unknown.

\textbf{Upperbound in Learned Geometry}: To understand the performance loss in the absence of perfect geometry i.e., a perfect CAD model, we evaluate our approach assuming that a perfect geometry,  texture-less CAD model, is available during training. Since we assume that we have a CAD model, we skip stage 1 in our training pipeline where geometry is learned. In the second stage of our approach, the density values are queried from the Density MLP learned in the first stage. In the presence of the CAD model, we skip the training of the first stage and train the second stage by querying density values directly from the CAD model. We refer to this experiment as Ours-G in Table \ref{table:tless}. We observe that there is no loss in performance between Ours and Ours-G. This shows that geometry from NeRF is already at the upper bound in terms of geometry extraction.
\begin{table}[!htp]\centering
\caption{Ablation studies on three objects on LM and LMO. We replace our NeRF with SoftRas based differentiable rendering.  Stages refer to the number of training stages used during training.}\label{tab: ablation2}
\scriptsize
\begin{tabular}{lrrrrrrr}\toprule
&SoftRas &Ours &Ours &SoftRas &Ours &Ours \\\midrule
Dataset &LM &LM &LM &LMO &LMO &LMO \\\midrule
Stages &2 &2 &1 &2 &2 &1 \\\midrule
Can &99.3 &99.3 &66.8 &82.2 &85.6 &50.1 \\
Driller &98.2 &98.3 &85.0 &63.9 &65.0 &38.3 \\
Iron box &99.8 &99.8 &94.6 &- &- &- \\\midrule
ADD &99.1 &99.1 &82.1 &73 &75.3 &44.2 \\
\bottomrule
\end{tabular}
\end{table}

\textbf{Two Stage vs Single Stage Training}:
Our pipeline comprises two stages. In the first stage, we train the NeRF to learn geometry. In the second stage, we freeze the geometry of NeRF and learn the feature space between CNN and NeRF. We perform an ablation to justify our two-stage training. In Table \ref{tab: ablation2}, we see that the performance of a two-stage pipeline is better than a single-stage.  This happens because feature learning gets intertwined with geometry learning leading to subpar performance if they are trained together. 


\textbf{PnP Ransac for Symmetric objects}
During inference, 2D-3D correspondences are estimated by finding the closest features between 3D point cloud features and 2D image features. We apply PnP ransac to estimate the final pose. Ideally, this should not work effectively for symmetric objects as the 2D correspondences could match to 3D correspondences in different symmetric configurations. However, as shown in Figure \ref{fig:CorrVis}, we observe that the correspondences are biased towards one symmetric configuration instead of distributing correspondences across multiple symmetric configurations. This is the reason for getting better performance despite using only PnP ransac which makes the approach much faster. We believe that this happens because of the way contrastive learning is formulated in our scenario. In our scenario, for symmetric objects, we sample negative samples uniformly from around the object which will definitely contain some positive samples from another symmetric configuration. Besides, a training batch could contain images from other symmetric configurations. The network chooses to prioritize one symmetric configuration over others based on the negative samples and the images in the batch. Because of this, the network might be preferring one symmetric configuration over others. So, the naive PnP ransac is enough to estimate the final 6d pose which makes the inference faster. This happens in the scenario where the batch of training images contains only one object. SurfEmb trains all the objects together and the batch contains different objects during each iteration which makes the SurfEmb learn features that spread correspondences similarly for all symmetric configurations. They had to employ a render and compare based intensive computational inference step to finalize the best hypothesis. Our inference takes 35ms which is much faster than SurfEmb which takes 1.2 secs for inference.    



\section{Conclusion}
We propose a novel pipeline combining Nerf and CNN to perform object pose estimation using weakly labeled data without requiring a cad model. The proposed feature learning approach enables us to handle symmetric objects by enforcing 3D and symmetry constraints and also facilitates faster inference. We achieve benchmark accuracy among approaches using only real images and relative pose labels on LM, LMO, T-Less datasets. Rotation estimation is robust with the pipeline which is evident from much improved results on LM and LM-Occlusion datasets in RGB-D setting. We propose a pose estimation pipeline that enables real training data based approach easier by working with weaker labels compared to the complex setup required to annotate 6D pose datasets without requiring a CAD model.   


\paragraph{Acknowledgements}
This work was partially funded by the German BMWK
under grant GEMIMEG-II-01MT20001A.
We present some additional ablations, qualitative and quantitative results in the following sections.
\section{Noisy Initial Pose Refinement}
In our pipeline, we propose an approach assuming that relative pose labels and bounding box labels are available for training. However, in reality, the relative poses acquired are pretty noisy. To demonstrate the applicability of our approach in real life, we perform an ablation with noisy initial relative pose labels. We preprocess the relative pose labels and refine them using Bundle Adjusting Radiance Fields, BARF\cite{lin2021barf}. We employ BARF to refine the noisy initial pose labels. BARF optimizes initial poses and networks together while training a NeRF. We employ BARF to refine our noisy initial poses and use the refined poses to train our pipeline.

We tweak the ground truth pose labels by adding random noise of 15 degrees on rotation matrices and 10 percent of the object's diameter as noise on the translation component. We present results in Table \ref{tab: ablationBarf} by running our pipeline with noisy poses and refined poses initialized with noisy poses. We observe that the accuracy with refined poses generated using BARF achieved 93.8\% accuracy which shows that employing BARF can handle the errors in initial pose labels. Although there is a performance drop(4.9\%) compared to ground truth poses, employing BARF improves the performance(23.3\%) compared to training with noisy poses. This demonstrates that our approach can be easily deployed in real life even in the presence of noisy pose labels. 

\begin{table}[!htp]\centering
\caption{Ablations: We perform an ablation with noisy initial pose estimates. We randomly perturb rotations with upto 15-degree error and translation with 10 percent of the diameter of the object on 3 objects, Can, Driller, and Iron from LineMOD Dataset. Labels refer to the pose labels used for training our pipeline. Noisy refers to our approach trained using noisy initial poses. BARF refers to applying training BARF on noisy pose labels to obtain refined poses. GT refers to our approach using ground truth pose labels. }\label{tab: ablationBarf}
\scriptsize
\begin{tabular}{crrr}\toprule

Labels  &Noisy &Noisy &GT \\\midrule\centering
Refinement &- &BARF &-  \\\midrule

Can  &69.9 &96.2 &99.0 \\
Driller  &65.7 &87.9 &98.1 \\
Iron  &76.0 &97.5 &99.0 \\\midrule
Mean  &70.5 &93.8 &98.7 \\
\bottomrule
\end{tabular}
\end{table}

\section{Faster Inference For Symmetric Objects}
\begin{figure*}[t]

     \centering
     \begin{subfigure}{.33\linewidth}
     \includegraphics[width=0.9\textwidth]{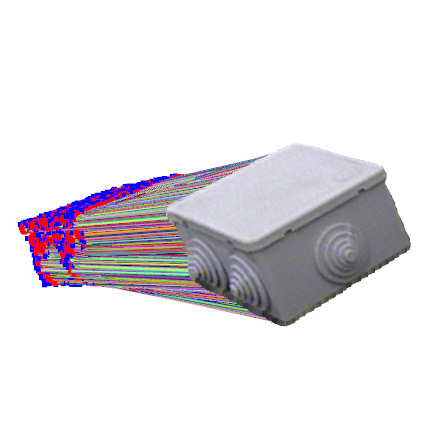}
      \caption{SurfEmb:View 1}
     \end{subfigure}%
     \begin{subfigure}{.33\linewidth}
     \centering
     \includegraphics[width=0.9\textwidth]{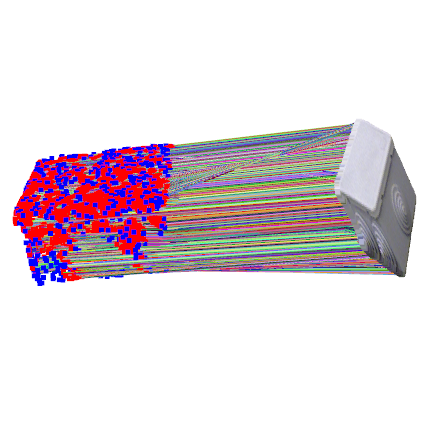}
     \caption{SurfEmb:View 2}
     \end{subfigure}%
     \hspace{0.023in}
     \begin{subfigure}{.33\linewidth}
     \centering
     \includegraphics[width=0.9\textwidth]{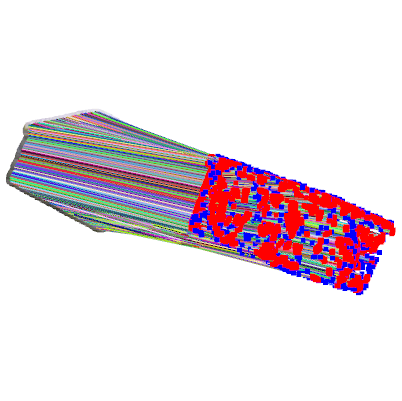}
     \caption{SurfEmb:View 2}
     \end{subfigure}
     \\ \vspace{0.1in}
     \centering
     \begin{subfigure}{.33\linewidth}
     \includegraphics[width=0.9\textwidth]{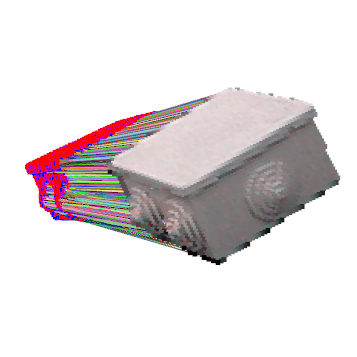}
     \caption{Ours:View 1}
     \end{subfigure}%
     \begin{subfigure}{.33\linewidth}
     \centering
     \includegraphics[width=0.9\textwidth]{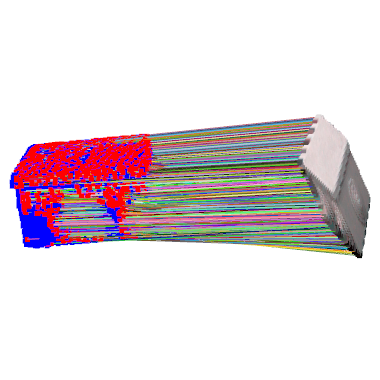}
     \caption{Ours:View 2}
    \end{subfigure}%
     \hspace{0.023in}
     \begin{subfigure}{.33\linewidth}
     \centering
     \includegraphics[width=0.9\textwidth]{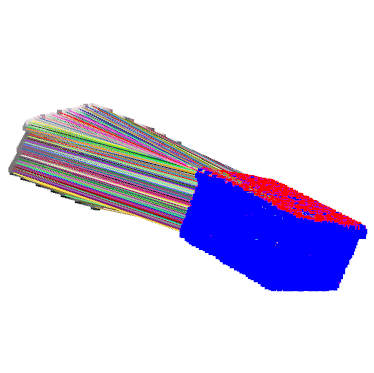}
         \caption{Ours:View 3}
     \end{subfigure}

     \caption{Visualization of correspondences for object 28 in T-Less: Discrete symmetric object. The visualization shows the 2D-3D correspondences between 2D masked pixels and the 3D point cloud of the object. We join lines between matched 2D-3D correspondences. The segmented 2D image points are indicated with their RGB color from the image. The blue pointcloud indicates the full point cloud of the object. The red point cloud indicates the correspondence to the current image(indicated with masked pixels). We visualize the correspondences in different views to show where the 3D correspondences on the object are matched.  The first row indicates the correspondence visualization for SurfEmb. The second row indicates the correspondences visualization of our approach. In SurfEmb, the correspondences are distributed around the object for the symmetric object. In our approach, the correspondences are biased towards only one symmetric configuration}
     \label{fig:symVis1}
\end{figure*}

\begin{figure*}[t]

     \centering
     \begin{subfigure}{.33\linewidth}
     \includegraphics[width=0.9\textwidth]{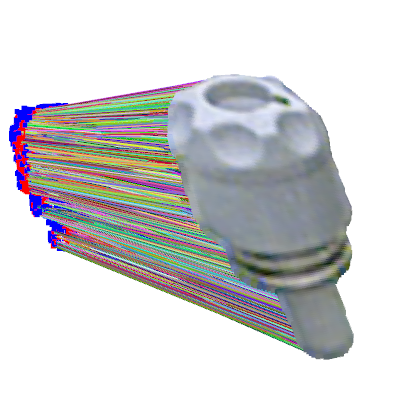}
      \caption{SurfEmb:View 1}
     \end{subfigure}%
     \begin{subfigure}{.33\linewidth}
     \centering
     \includegraphics[width=0.9\textwidth]{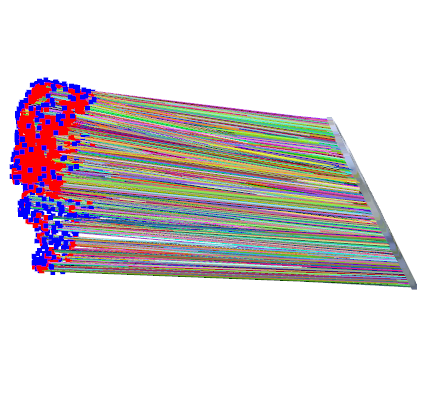}
     \caption{SurfEmb:View 2}
     \end{subfigure}%
     \hspace{0.023in}
     \begin{subfigure}{.33\linewidth}
     \centering
     \includegraphics[width=0.9\textwidth]{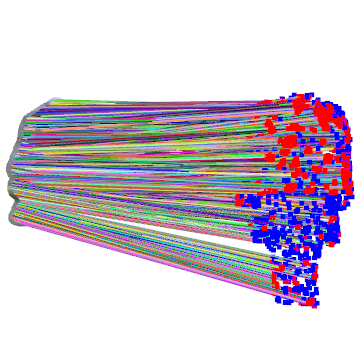}
     \caption{SurfEmb:View 2}
     \end{subfigure}
     \\ \vspace{0.1in}
     \centering
     \begin{subfigure}{.33\linewidth}
     \includegraphics[width=0.9\textwidth]{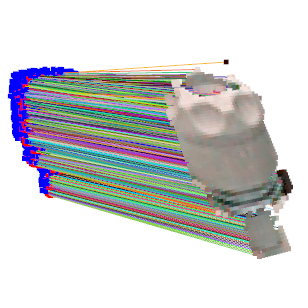}
     \caption{Ours:View 1}
     \end{subfigure}%
     \begin{subfigure}{.33\linewidth}
     \centering
     \includegraphics[width=0.9\textwidth]{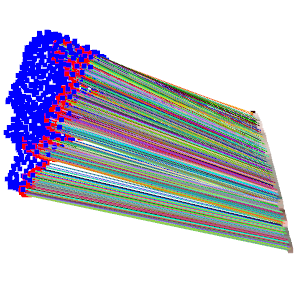}
     \caption{Ours:View 2}
    \end{subfigure}%
     \hspace{0.023in}
     \begin{subfigure}{.33\linewidth}
     \centering
     \includegraphics[width=0.9\textwidth]{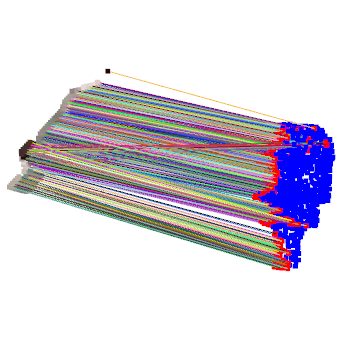}
         \caption{Ours:View 3}
     \end{subfigure}

     \caption{Visualization of correspondences for object 1 in T-Less: Continuous symmetric object. The visualization shows the 2D-3D correspondences between 2D masked pixels and the 3D point cloud of the object. We join lines between matched 2D-3D correspondences. The masked 2D image points are indicated by their color from the image. The blue point cloud indicates the full point cloud of the object. The red point cloud indicates the correspondences to the current image(indicated with masked pixels). We visualize the correspondences in different views to show where the 3D correspondences on the object are matched.  The first row indicates the correspondence visualization for SurfEmb. The second row indicates the corresponding visualization of our approach. In SurfEmb, the correspondences are distributed around the object for the symmetric object. In our approach, the correspondences are biased towards only one symmetric configuration}
     \label{fig:symVis2}
\end{figure*}

In the ablation section of the main paper, we show that the naive PnP Ransac is enough to estimate 6D pose even for symmetric objects. Although the features look similar visually as shown in \ref{fig:TLess1} and \ref{fig:TLess2}, there is still some discrepancy numerically between features which should have been the exact same value obeying their symmetries. This discrepancy leads to a bias towards one symmetric configuration when matched with a 2D image from CNN. SurfEmb trains the pipeline with all the objects comprising different objects in each batch with batch size 16. As they train with 30 objects, their batch contains different objects at each iteration. In our approach, a single batch contains 16 images of the same object. In Figures  \ref{fig:symVis1} and \ref{fig:symVis2}, we visualize the correspondences and compare how our correspondences are biased towards one symmetric configuration compared to SurfEmb.

In our approach, we train only one object at a time with a batch size of 16. We perform an ablation on how PnP Ransac is well suited even for symmetric objects in Table \ref{tab:tlessablation}. We perform ablations on a continuous symmetric object,  object 1,  in T-Less. We observe that the network doesn't perform well with batch size 1. So, a higher batch size is necessary for better performance. When the model is trained on one object, PnP Ransac which is faster also performs better(0.7\%) than SurfEmb's intensive evaluation pipeline. When the model is trained on 30 objects, PnP Ransac performs worse(6.1\%) compared to SurfEmb's evaluation. This shows that the correspondences are distributed around the object for symmetric objects when trained with 30 objects(a batch containing the utmost 1 image per object). By employing a higher batch size for a single object, we are able to achieve correspondences on the 3D model which are biased to one symmetric configuration for symmetric objects. This enables the handling of symmetric objects much easier compared to SurfEmb. For symmetric objects, SurfEmb needs an intense render and compare approach to find the best hypothesis as their correspondences are distributed around the object which makes it less suited to apply PnP Ransac. We can achieve comparable accuracy with faster inference time without any special handling for symmetric objects. Even though we managed to make the inference faster for symmetric objects, it comes at the cost of training a single model per object. So, our approach is still useful if real-time performance is desired at the cost of having a separate model per object. Also, we do not assume that we know about the object's symmetry before training and inference. We can handle symmetric objects despite not knowing about them being symmetric or not.  
\begin{table}[!htp]\centering
\caption{We perform an ablation with different batch sizes and different numbers of training objects per model. We evaluate the performance of the first object in T-Less which is Continuous Symmetric. Batch Size refers to the number of images used per batch during training. Training Objects refers to the number of objects trained in the model. PnP refers to naive PnP Ransac used to evaluate 6D pose from 2D-3D correspondences. SE refers to the evaluation pipeline proposed by SurfEmb which samples 10000 pose hypotheses and finds the best hypothesis by using intensive render and compare. PnP Ransac is much faster than  SurfEmb's evaluation. Ransac is clearly suitable even for symmetric objects when only one object is trained with a higher batch size. }\label{tab:tlessablation}
\scriptsize
\begin{tabular}{l|rrrrrr}\toprule

Experiment &Exp1&Exp2&Exp3&Exp4&Exp5\\\midrule
Batch Size &1 &16 &16&16&16  \\\midrule
Training Objects &1 &1 &1&30&30  \\\midrule
Evaluation & PnP &PnP&SE&PnP&SE\\\midrule

Mean  &3.0 &47.8 &47.1 & 34.9 & 41.0 \\
\bottomrule
\end{tabular}
\end{table}

\section{Training Details}
\begin{figure}
    \centering
    \includegraphics[width=60mm,scale=1]{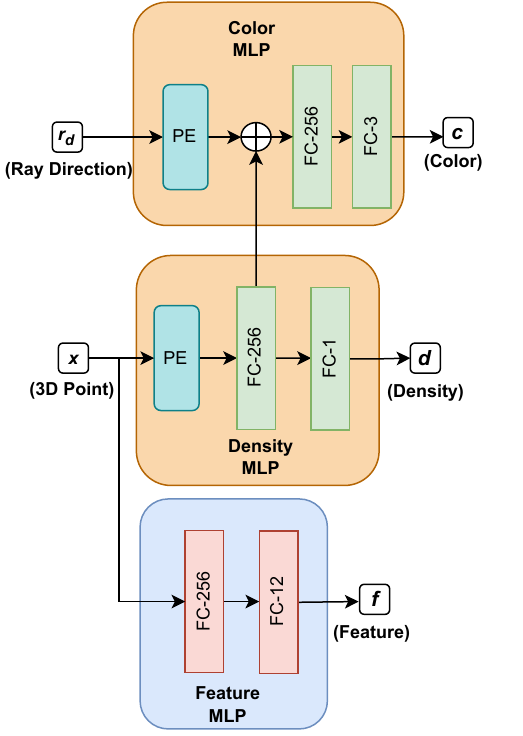}
    \caption{\textbf{NeRF Architecture}: Our NeRF comprises three MLPs, Density MLP, Color MLP and Feature MLP.  Density MLP takes in a discrete 3D point, $x$, and passes through the Position Encoding (PE) layer and three Fully Connected (FC) layers to predict density, $d$.  The green blocks are fully connected layers with Softplus activation function. Color MLP takes in a concatenated input from Ray Direction, $r_{d}$, and an intermediate feature from the second layer in Density MLP to predict color, $c$.  
    Feature MLP takes in a discrete 3D point, $x$, and passed through two fully connected layers to predict feature value, $f$. The pink blocks in Feature MLP employ Sine activation functions as proposed in Siren \cite{siren}. The orange blocks are trained during Stage 1 and the blue block along with CNN is trained during Stage 2.}
    \label{fig:nerfArch}
\end{figure}

We present some additional training details and network architecture details in this section. We employ a U-Net\cite{unet} based CNN and an adapted version of NeRF \cite{Nerf} to train our pipeline. Our Nerf consists of 3 MLPs, Density MLP, Color MLP, and Feature MLP. The architecture is depicted in Figure \ref{fig:nerfArch}.
\paragraph{Density MLP}
Density MLP takes in a single 3D coordinate and predicts density value. We use positional encoding for encoding 3D points with 60 harmonic functions as specified in Nerf\cite{Nerf}. We employ an MLP with three fully connected layers with Softplus activation. A 3D point is transformed into 360-dimensional positional encoding using 60 harmonic functions. The intermediate layers in MLP have 256-dimensional output and the final layer outputs a 1-dimensional density value.  

\paragraph{Color MLP}Color MLP takes in the ray direction and intermediate feature vector from the second layer of Density MLP as input to predict 3-dimensional color value. The three-dimensional ray direction vector is passed through the same position encoding as Density MLP to encode the 3D vector to create 360-dimensional output. The position-encoded ray direction vector and intermediate feature vector from Density MLP are concatenated together to create input for Color MLP. Color MLP is a two-layered MLP that employs the Softplus activation function after linear layers. The intermediate layer has 256-dimensional output.

\paragraph{Feature MLP}
Feature MLP takes in a 3D coordinate and predicts a 12-dimensional feature vector using a 2-layer MLP. We employ Siren MLP similar to SurfEmb\cite{surfemb}. As position encoding is already present in the intermediate layers in Siren\cite{siren} MLP implicitly, we do not use positional encoding to transform 3D coordinates. The intermediate layer has 256-dimensional output.
\paragraph{CNN}
We employ a U-Net\cite{unet} based CNN to predict feature images. The CNN takes in an image with the resolution, $224 \times 224 \times 3$ to predict the feature image of the resolution, $224 \times 224 \times 12$. We predict 12-dimensional feature vectors for each pixel. 
\subsection{Data Preparation}
We assume that the bounding box labels and relative pose labels between images are available for training. We use Segment Anything to segment the objects using the bounding box as input. We use segmentation masks, given images, and relative pose labels to train our pipeline. We use object crops and scale them to fit into $190 \times 190$ resolution while preserving the aspect ratio. To create synthetic occlusions with some part of the object removed, we create occlusion in the crops by randomly transforming the object out of the image and applying the inverse transformation so that the part of the image outside the image after the initial transformation is lost which gives us an image with some part of the object removed. We also remove some rectangular regions randomly from the object. We place the object crop in $224 \times 224$ resolution image at a random location and apply random scale, and rotations on the new image to create more augmentations. We also apply filters from Albumentations \cite{Albumentations} to add noise, contrast, brightness, and blur augmentations.  We follow the same procedure to create an image of size $224 \times 224$ from the object crop for inference. We do not perform augmentations on the inference image.
\subsection{Training}
We use an Nvidia GeForce GV102 GPU  with 12GB memory to train our models. Training Stage 1 takes 8 hours and training Stage 2 takes 6 hours. Inference time for a test sample takes 35 milliseconds. We train NeRF in stage 1 for 60K iterations with a batch size of 3. We use a learning rate of 1e-3. We use the NeRF implementation with ray sampler and renderer from Pytorch3d \cite{pytorch3d}. We sample 650 rays for an image with 512 points per ray during stage 1 as we need to learn geometry. In stage 2, we use stratified sampling to sample points only near the surface. We initially find the blend weights using the ray sampler with 512 points. We perform stratified sampling using blend weights to sample 8 points along the ray based on blend weights. Stratified sampling ensures that points very close to the surface are sampled.  We use only samples from stratified sampling to perform feature rendering in stage 2. We train Feature MLP and CNN during stage 2 using a learning rate of 3e-5 and 3e-4 respectively for 60K iterations with a batch size of 16. Instead of formulating the loss over the entire feature image, we sample 1024 samples from the feature image from CNN and 1024 samples at the same pixel locations in the feature image from NeRF. We sample 1024 negative samples by shooting rays from all the available cameras to get uniformly distributed negative samples across the object. This is done for each batch sample. So, a batch sample contains 1024 positive samples from both NeRF feature image and CNN feature image and 1024 negative samples. All the samples are 12-dimensional feature vectors. The contrastive loss, $L_{c}$, is formulated for 1024 pixels over 12-dimensional feature vectors from CNN feature image, $F$, NeRF Feature Image, $G$ and 1024 negative samples, $Z$, as follows:

\begin{equation}
    L_{c}=-\sum_{i=1}^{1024} \log \frac{\exp(F^{i} G^{i})}{\sum_{j=1}^{1024}\exp(F^{i} Z^{j})}
\end{equation}

$F_{i}$ and $G_{i}$ are 12-dimensional feature vectors from the CNN feature image,$ F$,  and NeRF image, $G$, at pixel $i$. $Z_{j}$ is a $j^{th}$12 dimensional feature vector from  1024 negative samples. 
\subsection{Learning Consistent Features}
We extract positive samples from NeRF feature image and CNN feature image to make sure that they predict the same features at the same pixel for a specific pose. We sample negative samples from different poses randomly to ensure that the negative samples are uniformly distributed across the shape. In the case of asymmetric objects, this helps to ensure that learned features are distinguishable based on their surface location on the object. In the case of symmetric objects, multiple points in symmetric configurations will map to similar features and the negative samples ensure that some form of discriminative feature space is learned while still obeying symmetry.

\subsection{Inference}
\paragraph{Point cloud extraction}
Before the inference phase, we extract the point cloud from the NeRF network by shooting rays from different viewpoints in training data and estimating the surface point and its corresponding feature. We extract the point cloud by estimating the point with blend weights along the ray and extracting the point corresponding to the maximum blend weights as the surface point. We can query per-point features for surface points by performing a forward pass for each 3D point through Feature MLP. So, we extract a point cloud with a per-point feature from our NeRF which is used for inference to establish correspondences.  

\paragraph{Inference with Image}
During inference, we preprocess our RGB image and pass it through CNN to extract the corresponding feature image. We also estimate segmentation mask using DpodV2. We extract features inside the segmentation mask from the feature image using the segmentation mask. We use these features to find correspondences with our point cloud with features. We compare each feature vector in the feature image with all feature vectors in point clouds by taking the dot product. We extract a match as the point corresponding to the feature with the maximum dot product score. After the 2D-3D correspondences are estimated, we simply use PnP Ransac with just 500 iterations to estimate the final 6D pose.   
\section{Limitations}
 The translation estimation can still be improved by decoupling it from rotation estimation. Our approach lags behind approaches using data from both CAD and real data. This needs to be addressed by a potential future work on Nerf to render realistic occlusions from real data. We need relative pose labels and a completely unsupervised approach without these labels is desired. 
 
\section{Additional Results}
\subsection{Linemod Test Results}
We add object-wise test results for Linemod Dataset and also extend the ablation studies to all objects. We use Segment anything masks and evaluate all the objects and observe a drop in accuracy of $3\%$. We also add results on all objects using 50 sample views for training instead of employing $~180$ views using in general setup. We improved our evaluation by tuning some hyperparameters in the inference pipeline and observed an increase in accuracy of $0.4\%$ compared to the results presented in the main paper.  Additional results are presented in Table \ref{tab:linemod }.

\begin{table*}[!htp]\centering
\caption{Linemod Results: We present results using RGB, RGB-D data, and CAD model. NP(Mask) refers to NeRF-Pose with an evaluation scheme comparing masks during each ransac iteration. Ours(S) refers to results obtained by segmentation masks generated using Segment Anything \cite{sam} instead of using ground truth segmentation masks. Ours(50) refers to our approach trained using just 50 views instead of using $~180$ views in the general evaluation setup. Ours(D) refers to our results with translation adjustment along the z-axis using a sensor depth map during inference.}\label{tab:linemod }
\scriptsize
\begin{tabular}{lrrrrrrrrrr}\toprule
Object &DpodV2\cite{DpoDv2} &FFB-6D\cite{he2021ffb6d} &LieNet\cite{Lienet} &Cai\cite{RLLG} &NP(Mask)\cite{Nerfpose} &Ours(50) &Ours(S) &Ours &Ours(D) \\\midrule
CAD &\cmark &\cmark &\xmark &\xmark &\xmark &\xmark &\xmark &\xmark &\xmark \\\midrule
Data &RGB &RGB-D &RGB &RGB &RGB &RGB &RGB &RGB &RGB-D \\\midrule
Ape &80 &98.4 &38.8 &52.9 &89.1 &55.2 &69.9 &71.5 &98.6 \\
BenchVice &99.7 &100 &71.2 &96.5 &99.3 &97.0 &99.3 &99.0 &100 \\
Camera &99.2 &99.9 &52.5 &87.8 &98.7 &84.0 &94.6 &96.3 &99.9 \\
Can &99.6 &99.8 &86.1 &86.8 &99.1 &93.5 &99.6 &99.3 &100 \\
Cat &95.1 &99.9 &66.2 &67.3 &97.1 &71.2 &73.3 &94.8 &100 \\
Drill &98.9 &100 &82.3 &88.7 &97.4 &95.5 &98.6 &98.3 &100 \\
Duck &79.5 &98.4 &32.5 &54.7 &90.3 &52.0 &63.7 &81.4 &99.5 \\
EggBox &99.6 &100 &79.4 &94.7 &99.6 &99.1 &100 &\textbf{100} &100 \\
Glue &99.8 &100 &63.7 &91.9 &98.1 &87.6 &95.0 &98.1 &99.9 \\
Holep &72.3 &99.8 &56.4 &75.4 &94.3 &67.0 &90.1 &93.5 &99.9 \\
Iron &99.4 &99.9 &65.1 &94.5 &98.1 &94.2 &99.5 &99.8 &100 \\
Lamp &96.3 &99.9 &89.4 &96.6 &97.9 &96.9 &98.7 &99.9 &99.9 \\
Phone &96.8 &99.7 &65.0 &89.2 &96.4 &88.0 &90.0 &93.9 &100 \\\midrule
Mean &93.5 &99.7 &65.2 &82.9 &96.6 &83.2 &90.2 &\textbf{94.3} &\textbf{99.8} \\
\bottomrule
\end{tabular}
\end{table*}

\subsection{T-Less Results}
We also add results with RGB-D data with the T-Less dataset to show that using depth in inference helps rectify translation. As we do not change rotation and adjust only translation, this shows that our method is robust in terms of rotation estimation. T-Less Dataset is licensed under a CC BY 4.0 license. Ours results are presented in Table \ref{table:tless}

\begin{table*}[!htp]\centering
\caption{T-Less Results: Evaluation on the T-Less dataset. NP refers to the NeRF-Pose approach. We also present AR on continuous symmetric(CS) and Discrete Symmetric(DS) objects on both NeRF-Pose(NP) and our approach. CAD refers to the models assuming the presence of a CAD model for training. Ours(D) refers to the results with translation adjusted using an input depth map from the sensor. }\label{tab:tless }
\scriptsize
\begin{tabular}{lrrrrrrrrrrr}\toprule
Appr &DpodV2\cite{DpoDv2} &DpodV2 &SurfEmb\cite{surfemb} &NP\cite{Nerfpose} &NP(CS) &Ours(CS) &NP(DS) &Ours(DS)  &Ours & Ours(D) \\\midrule
CAD &\cmark &\xmark &\cmark &\xmark &\xmark &\xmark &\xmark &\xmark &\xmark &\xmark \\\midrule
VSD &0.57 &0.46 &0.5 &0.45 &0.378 &0.42 &0.51 &0.56  &0.5& 0.58 \\
MSSD &0.62 &0.49 &0.53 &0.49 &0.42 &0.462 &0.56 &0.6  &0.54 &0.66\\
MSPD &0.76 &0.59 &0.83 &0.66 &0.672 &0.693 &0.65 &0.69  &0.7& 0.69 \\\midrule
AR &0.65 &0.51 &0.62 &0.54 &0.48 &\textbf{0.52} &0.58 &\textbf{0.63}  &\textbf{0.58} &0.65 \\
\bottomrule
\end{tabular}
\label{table:tless}
\end{table*}

\begin{figure*}[t]

     \centering
     \begin{subfigure}{.33\linewidth}
     \includegraphics[width=0.9\textwidth]{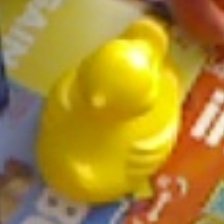}
     \end{subfigure}%
     \begin{subfigure}{.33\linewidth}
     \centering
     \includegraphics[width=0.9\textwidth]{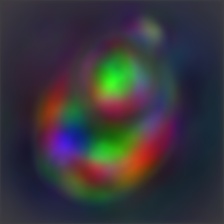}
     \end{subfigure}%
     \hspace{0.023in}
     \begin{subfigure}{.33\linewidth}
     \centering
     \includegraphics[width=0.9\textwidth]{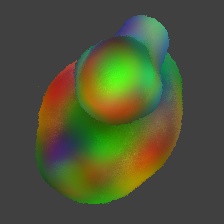}
     \end{subfigure}
     \\ \vspace{0.1in}
     \centering
     \begin{subfigure}{.33\linewidth}
     \includegraphics[width=0.9\textwidth]{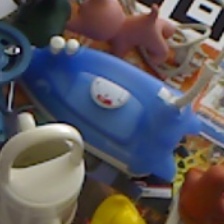}
     \end{subfigure}%
     \begin{subfigure}{.33\linewidth}
     \centering
     \includegraphics[width=0.9\textwidth]{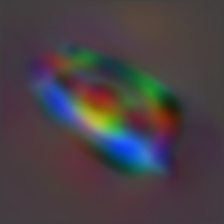}
     \end{subfigure}%
     \hspace{0.023in}
     \begin{subfigure}{.33\linewidth}
     \centering
     \includegraphics[width=0.9\textwidth]{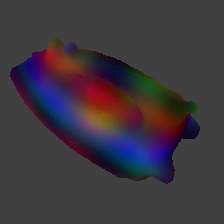}
     \end{subfigure}

     \vspace{0.1in}
     \centering
     \begin{subfigure}{.33\linewidth}
     \includegraphics[width=0.9\textwidth]{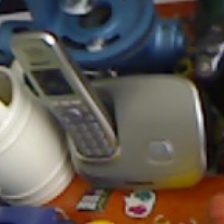}
     \end{subfigure}%
     \begin{subfigure}{.33\linewidth}
     \centering
     \includegraphics[width=0.9\textwidth]{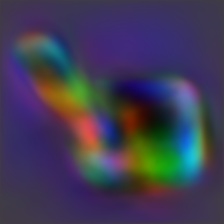}
     \end{subfigure}%
     \hspace{0.023in}
     \begin{subfigure}{.33\linewidth}
     \centering
     \includegraphics[width=0.9\textwidth]{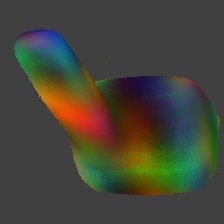}
     \end{subfigure}

     \vspace{0.1in}
     \centering
     \begin{subfigure}{.33\linewidth}
     \includegraphics[width=0.9\textwidth]{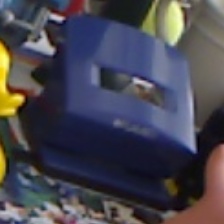}
     \end{subfigure}%
     \begin{subfigure}{.33\linewidth}
     \centering
     \includegraphics[width=0.9\textwidth]{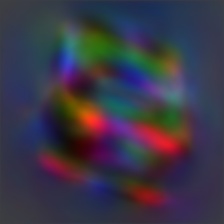}
     \end{subfigure}%
     \hspace{0.023in}
     \begin{subfigure}{.33\linewidth}
     \centering
     \includegraphics[width=0.9\textwidth]{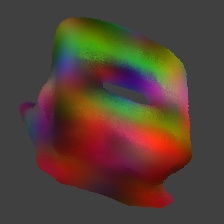}
     \end{subfigure}
     %

     \caption{Visualization of learned feature representation of Linemod objects. The first image in each row is the input RGB image and the second image is the output feature image from CNN and the third image is the feature image rendered from NeRF at the given pose. Visualizations are from objects, Duck, Ironbox, Phone and Holepunch in Linemod dataset.}
     \label{fig:Linemod}
\end{figure*}

\begin{figure*}[ht]

     \centering
     \begin{subfigure}{.33\linewidth}
     \includegraphics[width=0.9\textwidth]{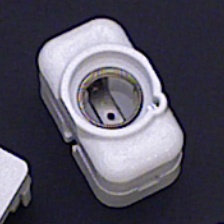}
     \end{subfigure}%
     \begin{subfigure}{.33\linewidth}
     \centering
     \includegraphics[width=0.9\textwidth]{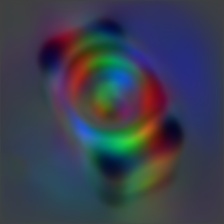}
     \end{subfigure}%
     \hspace{0.023in}
     \begin{subfigure}{.33\linewidth}
     \centering
     \includegraphics[width=0.9\textwidth]{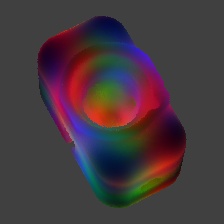}
     \end{subfigure}
     \\ \vspace{0.1in}
     \centering
     \begin{subfigure}{.33\linewidth}
     \includegraphics[width=0.9\textwidth]{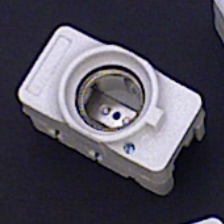}
     \end{subfigure}%
     \begin{subfigure}{.33\linewidth}
     \centering
     \includegraphics[width=0.9\textwidth]{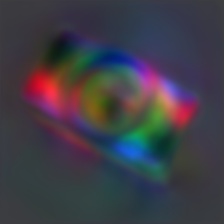}
     \end{subfigure}%
     \hspace{0.023in}
     \begin{subfigure}{.33\linewidth}
     \centering
     \includegraphics[width=0.9\textwidth]{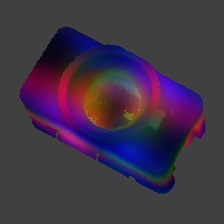}
     \end{subfigure}

     \vspace{0.1in}
     \centering
     \begin{subfigure}{.33\linewidth}
     \includegraphics[width=0.9\textwidth]{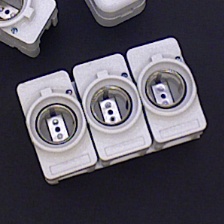}
     \end{subfigure}%
     \begin{subfigure}{.33\linewidth}
     \centering
     \includegraphics[width=0.9\textwidth]{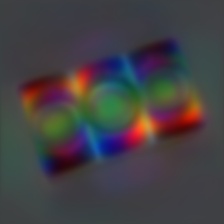}
     \end{subfigure}%
     \hspace{0.023in}
     \begin{subfigure}{.33\linewidth}
     \centering
     \includegraphics[width=0.9\textwidth]{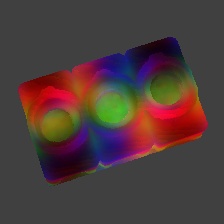}
     \end{subfigure}

     \vspace{0.1in}
     \centering
     \begin{subfigure}{.33\linewidth}
     \includegraphics[width=0.9\textwidth]{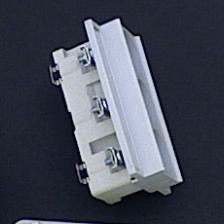}
     \end{subfigure}%
     \begin{subfigure}{.33\linewidth}
     \centering
     \includegraphics[width=0.9\textwidth]{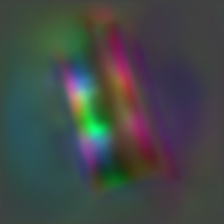}
     \end{subfigure}%
     \hspace{0.023in}
     \begin{subfigure}{.33\linewidth}
     \centering
     \includegraphics[width=0.9\textwidth]{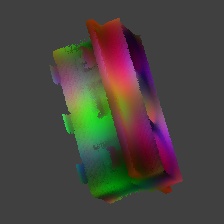}
     \end{subfigure}

     \caption{Visualization of learned feature representation of Discrete Symmetric objects in T-Less objects. The first image in each row is the input RGB image and the second image is the output feature image from CNN and the third image is the feature image rendered from NeRF at the given pose. Visualizations are from objects 5, 6, 7, 9 in T-Less dataset.}
     \label{fig:TLess1}
\end{figure*}

\begin{figure*}[ht]

     \centering
     \begin{subfigure}{.33\linewidth}
     \includegraphics[width=0.9\textwidth]{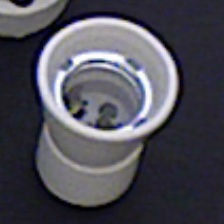}
     \end{subfigure}%
     \begin{subfigure}{.33\linewidth}
     \centering
     \includegraphics[width=0.9\textwidth]{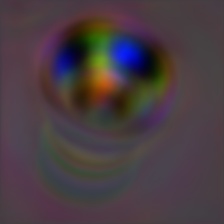}
     \end{subfigure}%
     \hspace{0.023in}
     \begin{subfigure}{.33\linewidth}
     \centering
     \includegraphics[width=0.9\textwidth]{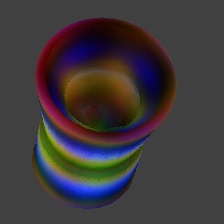}
     \end{subfigure}
     \\ \vspace{0.1in}
     \centering
     \begin{subfigure}{.33\linewidth}
     \includegraphics[width=0.9\textwidth]{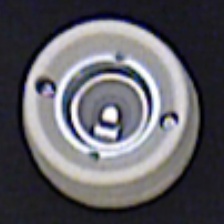}
     \end{subfigure}%
     \begin{subfigure}{.33\linewidth}
     \centering
     \includegraphics[width=0.9\textwidth]{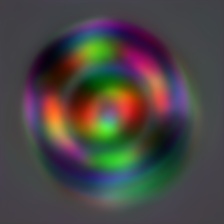}
     \end{subfigure}%
     \hspace{0.023in}
     \begin{subfigure}{.33\linewidth}
     \centering
     \includegraphics[width=0.9\textwidth]{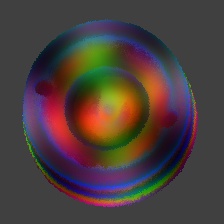}
     \end{subfigure}

     \vspace{0.1in}
     \centering
     \begin{subfigure}{.33\linewidth}
     \includegraphics[width=0.9\textwidth]{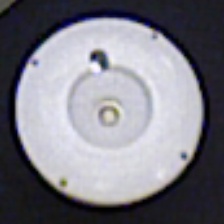}
     \end{subfigure}%
     \begin{subfigure}{.33\linewidth}
     \centering
     \includegraphics[width=0.9\textwidth]{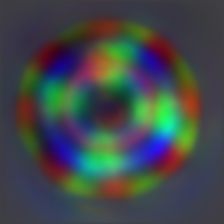}
     \end{subfigure}%
     \hspace{0.023in}
     \begin{subfigure}{.33\linewidth}
     \centering
     \includegraphics[width=0.9\textwidth]{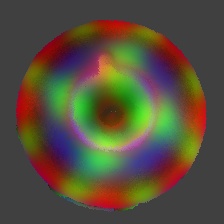}
     \end{subfigure}

     \vspace{0.1in}
     \centering
     \begin{subfigure}{.33\linewidth}
     \includegraphics[width=0.9\textwidth]{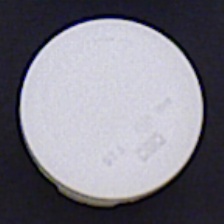}
     \end{subfigure}%
     \begin{subfigure}{.33\linewidth}
     \centering
     \includegraphics[width=0.9\textwidth]{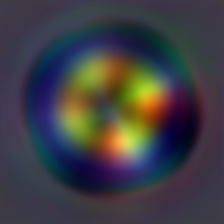}
     \end{subfigure}%
     \hspace{0.023in}
     \begin{subfigure}{.33\linewidth}
     \centering
     \includegraphics[width=0.9\textwidth]{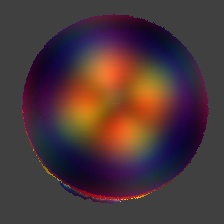}
     \end{subfigure}

     \caption{Visualization of learned feature representation of Continuous Symmetric objects in T-Less objects. The first image in each row is the input RGB image and the second image is the output feature image from CNN and the third image is the feature image rendered from NeRF at the given pose. Visualizations are from objects 14,16,3,30 in T-Less dataset.}
     \label{fig:TLess2}
\end{figure*}

\subsection{Qualitative Results}
We present some qualitative results and learned feature maps in NeRF renderings and predicted feature maps from CNN in Figures \ref{fig:Linemod}, \ref{fig:TLess1}, \ref{fig:TLess2} in this section. 
\clearpage
\clearpage


{
    \small
    \bibliographystyle{ieeenat_fullname}
    \bibliography{main}

\begin{thebibliography}{55}
\providecommand{\natexlab}[1]{#1}
\providecommand{\url}[1]{\texttt{#1}}
\expandafter\ifx\csname urlstyle\endcsname\relax
  \providecommand{\doi}[1]{doi: #1}\else
  \providecommand{\doi}{doi: \begingroup \urlstyle{rm}\Url}\fi

\bibitem[Arun et~al.(1987)Arun, Huang, and Blostein]{ICP}
K.~S. Arun, T.~S. Huang, and S.~D. Blostein.
\newblock Least-squares fitting of two 3-d point sets.
\newblock \emph{IEEE Transactions on Pattern Analysis and Machine Intelligence}, 1987.

\bibitem[Brachmann et~al.(2014)Brachmann, Krull, Michel, Gumhold, Shotton, and Rother]{occlusion}
Eric Brachmann, Alexander Krull, Frank Michel, Stefan Gumhold, Jamie Shotton, and Carsten Rother.
\newblock Learning 6d object pose estimation using 3d object coordinates.
\newblock In \emph{ECCV}, 2014.

\bibitem[Busam et~al.(2020)Busam, Jung, and Navab]{busam2020like}
Benjamin Busam, Hyun~Jun Jung, and Nassir Navab.
\newblock I like to move it: 6d pose estimation as an action decision process.
\newblock \emph{arXiv preprint arXiv:2009.12678}, 2020.

\bibitem[Buslaev et~al.(2020)Buslaev, Iglovikov, Khvedchenya, Parinov, Druzhinin, and Kalinin]{Albumentations}
Alexander Buslaev, Vladimir~I. Iglovikov, Eugene Khvedchenya, Alex Parinov, Mikhail Druzhinin, and Alexandr~A. Kalinin.
\newblock Albumentations: Fast and flexible image augmentations.
\newblock \emph{Information}, 11\penalty0 (2), 2020.

\bibitem[Cai et~al.(2022)Cai, Heikkil{\"a}, and Rahtu]{ove6d}
Dingding Cai, Janne Heikkil{\"a}, and Esa Rahtu.
\newblock Ove6d: Object viewpoint encoding for depth-based 6d object pose estimation.
\newblock In \emph{Proceedings of the IEEE/CVF Conference on Computer Vision and Pattern Recognition}, pages 6803--6813, 2022.

\bibitem[Cai and Reid(2020)]{RLLG}
Ming Cai and Ian Reid.
\newblock Reconstruct locally, localize globally: A model free method for object pose estimation.
\newblock In \emph{Proceedings of the IEEE/CVF Conference on Computer Vision and Pattern Recognition (CVPR)}, 2020.

\bibitem[Caron et~al.(2021)Caron, Touvron, Misra, J\'egou, Mairal, Bojanowski, and Joulin]{dino}
Mathilde Caron, Hugo Touvron, Ishan Misra, Herv\'e J\'egou, Julien Mairal, Piotr Bojanowski, and Armand Joulin.
\newblock Emerging properties in self-supervised vision transformers.
\newblock In \emph{Proceedings of the International Conference on Computer Vision (ICCV)}, 2021.

\bibitem[Chen et~al.(2023)Chen, Manhardt, Navab, and Busam]{texpose}
Hanzhi Chen, Fabian Manhardt, Nassir Navab, and Benjamin Busam.
\newblock Texpose: Neural texture learning for self-supervised 6d object pose estimation.
\newblock In \emph{IEEE/CVF Conference on Computer Vision and Pattern Recognition (CVPR)}, 2023.

\bibitem[Chen et~al.(2020)Chen, Jia, Chang, Duan, and Leonardis]{G2lNet}
Wei Chen, Xi Jia, Hyung~Jin Chang, Jinming Duan, and Ales Leonardis.
\newblock G2l-net: Global to local network for real-time 6d pose estimation with embedding vector features.
\newblock In \emph{IEEE/CVF Conference on Computer Vision and Pattern Recognition (CVPR)}, 2020.

\bibitem[Denninger et~al.(2019)Denninger, Sundermeyer, Winkelbauer, Zidan, Olefir, Elbadrawy, Lodhi, and Katam]{denninger2019blenderproc}
Maximilian Denninger, Martin Sundermeyer, Dominik Winkelbauer, Youssef Zidan, Dmitry Olefir, Mohamad Elbadrawy, Ahsan Lodhi, and Harinandan Katam.
\newblock Blenderproc.
\newblock \emph{arXiv preprint arXiv:1911.01911}, 2019.

\bibitem[Denninger et~al.(2023)Denninger, Winkelbauer, Sundermeyer, Boerdijk, Knauer, Strobl, Humt, and Triebel]{blenderProc}
Maximilian Denninger, Dominik Winkelbauer, Martin Sundermeyer, Wout Boerdijk, Markus Knauer, Klaus~H. Strobl, Matthias Humt, and Rudolph Triebel.
\newblock Blenderproc2: A procedural pipeline for photorealistic rendering.
\newblock \emph{Journal of Open Source Software}, 8\penalty0 (82):\penalty0 4901, 2023.

\bibitem[Fang et~al.(2020)Fang, Wang, Gou, and Lu]{fang2020graspnet}
Hao-Shu Fang, Chenxi Wang, Minghao Gou, and Cewu Lu.
\newblock Graspnet-1billion: A large-scale benchmark for general object grasping.
\newblock In \emph{Proceedings of the IEEE/CVF conference on computer vision and pattern recognition}, pages 11444--11453, 2020.

\bibitem[Haugaard and Buch(2022)]{surfemb}
R. Haugaard and A. Buch.
\newblock Surfemb: Dense and continuous correspondence distributions for object pose estimation with learnt surface embeddings.
\newblock In \emph{2022 IEEE/CVF Conference on Computer Vision and Pattern Recognition (CVPR)}, pages 6739--6748, Los Alamitos, CA, USA, 2022. IEEE Computer Society.

\bibitem[He et~al.(2022)He, Sun, Wang, Huang, Bao, and Zhou]{oneposeplusplus}
Xingyi He, Jiaming Sun, Yuang Wang, Di Huang, Hujun Bao, and Xiaowei Zhou.
\newblock Onepose++: Keypoint-free one-shot object pose estimation without {CAD} models.
\newblock In \emph{Advances in Neural Information Processing Systems}, 2022.

\bibitem[He et~al.(2020)He, Sun, Huang, Liu, Fan, and Sun]{he2020pvn3d}
Yisheng He, Wei Sun, Haibin Huang, Jianran Liu, Haoqiang Fan, and Jian Sun.
\newblock Pvn3d: A deep point-wise 3d keypoints voting network for 6dof pose estimation.
\newblock In \emph{CVPR}, 2020.

\bibitem[He et~al.(2021)He, Huang, Fan, Chen, and Sun]{he2021ffb6d}
Yisheng He, Haibin Huang, Haoqiang Fan, Qifeng Chen, and Jian Sun.
\newblock Ffb6d: A full flow bidirectional fusion network for 6d pose estimation.
\newblock In \emph{Proceedings of the IEEE/CVF Conference on Computer Vision and Pattern Recognition}, pages 3003--3013, 2021.

\bibitem[Hinterstoisser et~al.(2012)Hinterstoisser, Lepetit, Ilic, Holzer, Bradski, Konolige, , and Navab]{hinterstoisser2012accv}
S. Hinterstoisser, V. Lepetit, S. Ilic, S. Holzer, G. Bradski, K. Konolige, , and N. Navab.
\newblock Model based training, detection and pose estimation of texture-less 3d objects in heavily cluttered scenes.
\newblock In \emph{Asian Conference on Computer Vision}, 2012.

\bibitem[Hodan and Melenovsky(2019)]{bop}
Tomas Hodan and Antonin Melenovsky.
\newblock Bop: Benchmark for 6d object pose estimation: \url{https://bop.felk.cvut.cz/home/}, 2019.

\bibitem[Hoda{\v{n}} et~al.(2017)Hoda{\v{n}}, Haluza, Obdr{\v{z}}{\'a}lek, Matas, Lourakis, and Zabulis]{tless}
Tom{\'a}{\v{s}} Hoda{\v{n}}, Pavel Haluza, {\v{S}}t{\v{e}}p{\'a}n Obdr{\v{z}}{\'a}lek, Ji{\v{r}}{\'\i} Matas, Manolis Lourakis, and Xenophon Zabulis.
\newblock {T-LESS}: An {RGB-D} dataset for {6D} pose estimation of texture-less objects.
\newblock \emph{IEEE Winter Conference on Applications of Computer Vision (WACV)}, 2017.

\bibitem[Jung et~al.(2022)Jung, Wu, Ruhkamp, Schieber, Wang, Rizzoli, Zhao, Meier, Roth, Navab, et~al.]{jung2022housecat6d}
HyunJun Jung, Shun-Cheng Wu, Patrick Ruhkamp, Hannah Schieber, Pengyuan Wang, Giulia Rizzoli, Hongcheng Zhao, Sven~Damian Meier, Daniel Roth, Nassir Navab, et~al.
\newblock Housecat6d--a large-scale multi-modal category level 6d object pose dataset with household objects in realistic scenarios.
\newblock \emph{arXiv preprint arXiv:2212.10428}, 2022.

\bibitem[Karnati et~al.(2022)Karnati, Seal, Yazidi, and Krejcar]{Lienet}
Mohan Karnati, Ayan Seal, Anis Yazidi, and Ondrej Krejcar.
\newblock Lienet: A deep convolution neural network framework for detecting deception.
\newblock \emph{IEEE Transactions on Cognitive and Developmental Systems}, 14\penalty0 (3):\penalty0 971--984, 2022.

\bibitem[Kirillov et~al.(2023)Kirillov, Mintun, Ravi, Mao, Rolland, Gustafson, Xiao, Whitehead, Berg, Lo, Doll{\'a}r, and Girshick]{sam}
Alexander Kirillov, Eric Mintun, Nikhila Ravi, Hanzi Mao, Chloe Rolland, Laura Gustafson, Tete Xiao, Spencer Whitehead, Alexander~C. Berg, Wan-Yen Lo, Piotr Doll{\'a}r, and Ross Girshick.
\newblock Segment anything.
\newblock \emph{arXiv:2304.02643}, 2023.

\bibitem[Kobayashi et~al.(2022)Kobayashi, Matsumoto, and Sitzmann]{distilledfeaturefields}
Sosuke Kobayashi, Eiichi Matsumoto, and Vincent Sitzmann.
\newblock Decomposing nerf for editing via feature field distillation.
\newblock In \emph{Advances in Neural Information Processing Systems}, 2022.

\bibitem[{Labbe} et~al.(2020){Labbe}, {Carpentier}, {Aubry}, and {Sivic}]{cosypose}
Y. {Labbe}, J. {Carpentier}, M. {Aubry}, and J. {Sivic}.
\newblock Cosypose: Consistent multi-view multi-object 6d pose estimation.
\newblock In \emph{Proceedings of the European Conference on Computer Vision (ECCV)}, 2020.

\bibitem[Li et~al.(2023)Li, Vutukur, Yu, Shugurov, Busam, Yang, and Ilic]{Nerfpose}
Fu Li, Shishir~Reddy Vutukur, Hao Yu, Ivan Shugurov, Benjamin Busam, Shaowu Yang, and Slobodan Ilic.
\newblock Nerf-pose: A first-reconstruct-then-regress approach for weakly-supervised 6d object pose estimation.
\newblock In \emph{Proceedings of the IEEE/CVF International Conference on Computer Vision}, pages 2123--2133, 2023.

\bibitem[Lin et~al.(2021)Lin, Ma, Torralba, and Lucey]{lin2021barf}
Chen-Hsuan Lin, Wei-Chiu Ma, Antonio Torralba, and Simon Lucey.
\newblock Barf: Bundle-adjusting neural radiance fields.
\newblock In \emph{IEEE International Conference on Computer Vision ({ICCV})}, 2021.

\bibitem[Lin et~al.(2014)Lin, Maire, Belongie, Hays, Perona, Ramanan, Doll{\'a}r, and Zitnick]{lin2014microsoft}
Tsung-Yi Lin, Michael Maire, Serge Belongie, James Hays, Pietro Perona, Deva Ramanan, Piotr Doll{\'a}r, and C~Lawrence Zitnick.
\newblock Microsoft coco: Common objects in context.
\newblock In \emph{Computer Vision--ECCV 2014: 13th European Conference, Zurich, Switzerland, September 6-12, 2014, Proceedings, Part V 13}, pages 740--755. Springer, 2014.

\bibitem[Liu et~al.(2019)Liu, Li, Chen, and Li]{softras}
Shichen Liu, Tianye Li, Weikai Chen, and Hao Li.
\newblock Soft rasterizer: A differentiable renderer for image-based 3d reasoning.
\newblock \emph{The IEEE International Conference on Computer Vision (ICCV)}, 2019.

\bibitem[Marion et~al.(2018)Marion, Florence, Manuelli, and Tedrake]{marion2018label}
Pat Marion, Peter~R Florence, Lucas Manuelli, and Russ Tedrake.
\newblock Label fusion: A pipeline for generating ground truth labels for real rgbd data of cluttered scenes.
\newblock In \emph{2018 IEEE International Conference on Robotics and Automation (ICRA)}, pages 3235--3242. IEEE, 2018.

\bibitem[Mildenhall et~al.(2020)Mildenhall, Srinivasan, Tancik, Barron, Ramamoorthi, and Ng]{Nerf}
Ben Mildenhall, Pratul~P. Srinivasan, Matthew Tancik, Jonathan~T. Barron, Ravi Ramamoorthi, and Ren Ng.
\newblock Nerf: Representing scenes as neural radiance fields for view synthesis.
\newblock \emph{CoRR}, abs/2003.08934, 2020.

\bibitem[Neverova et~al.(2020)Neverova, Novotn{\'{y}}, Khalidov, Szafraniec, Labatut, and Vedaldi]{ContSurface}
Natalia Neverova, David Novotn{\'{y}}, Vasil Khalidov, Marc Szafraniec, Patrick Labatut, and Andrea Vedaldi.
\newblock Continuous surface embeddings.
\newblock \emph{CoRR}, abs/2011.12438, 2020.

\bibitem[Park et~al.(2019)Park, Patten, and Vincze]{pix2pose}
Kiru Park, Timothy Patten, and Markus Vincze.
\newblock Pix2pose: Pix2pose: Pixel-wise coordinate regression of objects for 6d pose estimation.
\newblock In \emph{The IEEE International Conference on Computer Vision (ICCV)}, 2019.

\bibitem[Park et~al.(2020)Park, Mousavian, Xiang, and Fox]{latentfusion}
Keunhong Park, Arsalan Mousavian, Yu Xiang, and Dieter Fox.
\newblock Latentfusion: End-to-end differentiable reconstruction and rendering for unseen object pose estimation.
\newblock In \emph{Proceedings of the IEEE Conference on Computer Vision and Pattern Recognition}, 2020.

\bibitem[Ravi et~al.(2020)Ravi, Reizenstein, Novotny, Gordon, Lo, Johnson, and Gkioxari]{pytorch3d}
Nikhila Ravi, Jeremy Reizenstein, David Novotny, Taylor Gordon, Wan-Yen Lo, Justin Johnson, and Georgia Gkioxari.
\newblock Accelerating 3d deep learning with pytorch3d.
\newblock \emph{arXiv:2007.08501}, 2020.

\bibitem[Ronneberger et~al.(2015)Ronneberger, Fischer, and Brox]{unet}
Olaf Ronneberger, Philipp Fischer, and Thomas Brox.
\newblock U-net: Convolutional networks for biomedical image segmentation, 2015.

\bibitem[Sch\"{o}nberger and Frahm(2016)]{colmap}
Johannes~Lutz Sch\"{o}nberger and Jan-Michael Frahm.
\newblock Structure-from-motion revisited.
\newblock In \emph{Conference on Computer Vision and Pattern Recognition (CVPR)}, 2016.

\bibitem[Shugurov et~al.(2021)Shugurov, Zakharov, and Ilic]{DpoDv2}
Ivan Shugurov, Sergey Zakharov, and Slobodan Ilic.
\newblock Dpodv2: Dense correspondence-based 6 dof pose estimation.
\newblock \emph{IEEE transactions on pattern analysis and machine intelligence}, PP, 2021.

\bibitem[Shugurov et~al.(2022)Shugurov, Li, Busam, and Ilic]{osop}
Ivan Shugurov, Fu Li, Benjamin Busam, and Slobodan Ilic.
\newblock Osop: A multi-stage one shot object pose estimation framework, 2022.

\bibitem[Sitzmann et~al.(2020)Sitzmann, Martel, Bergman, Lindell, and Wetzstein]{siren}
Vincent Sitzmann, Julien~N.P. Martel, Alexander~W. Bergman, David~B. Lindell, and Gordon Wetzstein.
\newblock Implicit neural representations with periodic activation functions.
\newblock In \emph{Proc. NeurIPS}, 2020.

\bibitem[Su et~al.(2022{\natexlab{a}})Su, Saleh, Fetzer, Rambach, Navab, Busam, Stricker, and Tombari]{su2022zebrapose}
Yongzhi Su, Mahdi Saleh, Torben Fetzer, Jason Rambach, Nassir Navab, Benjamin Busam, Didier Stricker, and Federico Tombari.
\newblock Zebrapose: Coarse to fine surface encoding for 6dof object pose estimation.
\newblock In \emph{Proceedings of the IEEE/CVF Conference on Computer Vision and Pattern Recognition}, pages 6738--6748, 2022{\natexlab{a}}.

\bibitem[Su et~al.(2022{\natexlab{b}})Su, Saleh, Fetzer, Rambach, Navab, Busam, Stricker, and Tombari]{zebrapose}
Yongzhi Su, Mahdi Saleh, Torben Fetzer, Jason Rambach, Nassir Navab, Benjamin Busam, Didier Stricker, and Federico Tombari.
\newblock Zebrapose: Coarse to fine surface encoding for 6dof object pose estimation.
\newblock \emph{arXiv preprint arXiv:2203.09418}, 2022{\natexlab{b}}.

\bibitem[Sun et~al.(2022)Sun, Wang, Zhang, He, Zhao, Zhang, and Zhou]{onepose}
Jiaming Sun, Zihao Wang, Siyu Zhang, Xingyi He, Hongcheng Zhao, Guofeng Zhang, and Xiaowei Zhou.
\newblock {OnePose}: One-shot object pose estimation without {CAD} models.
\newblock \emph{CVPR}, 2022.

\bibitem[Tschernezki et~al.(2022)Tschernezki, Laina, Larlus, and Vedaldi]{n3f}
Vadim Tschernezki, Iro Laina, Diane Larlus, and Andrea Vedaldi.
\newblock Neural feature fusion fields: {3D} distillation of self-supervised {2D} image representations.
\newblock In \emph{Proceedings of the International Conference on {3D} Vision (3DV)}, 2022.

\bibitem[van~den Oord et~al.(2019)van~den Oord, Li, and Vinyals]{infonce}
Aaron van~den Oord, Yazhe Li, and Oriol Vinyals.
\newblock Representation learning with contrastive predictive coding, 2019.

\bibitem[Vutukur et~al.(2022)Vutukur, Shugurov, Busam, Hutter, and Ilic]{welsa}
Shishir~Reddy Vutukur, Ivan Shugurov, Benjamin Busam, Andreas Hutter, and Slobodan Ilic.
\newblock Welsa: Learning to predict 6d pose from weakly labeled data using shape alignment.
\newblock In \emph{Computer Vision--ECCV 2022: 17th European Conference, Tel Aviv, Israel, October 23--27, 2022, Proceedings, Part VIII}, pages 645--661. Springer, 2022.

\bibitem[Wang et~al.(2019)Wang, Xu, Zhu, Mart{\'\i}n-Mart{\'\i}n, Lu, Fei-Fei, and Savarese]{densefusion}
Chen Wang, Danfei Xu, Yuke Zhu, Roberto Mart{\'\i}n-Mart{\'\i}n, Cewu Lu, Li Fei-Fei, and Silvio Savarese.
\newblock Densefusion: 6d object pose estimation by iterative dense fusion.
\newblock In \emph{Computer Vision and Pattern Recognition (CVPR)}, 2019.

\bibitem[Wang et~al.(2020)Wang, Manhardt, Shao, Ji, Navab, and Tombari]{self6d}
Gu Wang, Fabian Manhardt, Jianzhun Shao, Xiangyang Ji, Nassir Navab, and Federico Tombari.
\newblock Self6d: Self-supervised monocular 6d object pose estimation.
\newblock In \emph{Computer Vision -- ECCV 2020}, pages 108--125, Cham, 2020. Springer International Publishing.

\bibitem[Wang et~al.(2021{\natexlab{a}})Wang, Manhardt, Tombari, and Ji]{gdrn}
Gu Wang, Fabian Manhardt, Federico Tombari, and Xiangyang Ji.
\newblock {GDR-Net}: Geometry-guided direct regression network for monocular 6d object pose estimation.
\newblock In \emph{IEEE/CVF Conference on Computer Vision and Pattern Recognition (CVPR)}, pages 16611--16621, 2021{\natexlab{a}}.

\bibitem[Wang et~al.(2021{\natexlab{b}})Wang, Manhardt, Minciullo, Garattoni, Meier, Navab, and Busam]{wang2021demograsp}
Pengyuan Wang, Fabian Manhardt, Luca Minciullo, Lorenzo Garattoni, Sven Meier, Nassir Navab, and Benjamin Busam.
\newblock Demograsp: Few-shot learning for robotic grasping with human demonstration.
\newblock In \emph{2021 IEEE/RSJ International Conference on Intelligent Robots and Systems (IROS)}, pages 5733--5740. IEEE, 2021{\natexlab{b}}.

\bibitem[Wang et~al.(2022)Wang, Jung, Li, Shen, Srikanth, Garattoni, Meier, Navab, and Busam]{wang2022phocal}
Pengyuan Wang, HyunJun Jung, Yitong Li, Siyuan Shen, Rahul~Parthasarathy Srikanth, Lorenzo Garattoni, Sven Meier, Nassir Navab, and Benjamin Busam.
\newblock Phocal: A multi-modal dataset for category-level object pose estimation with photometrically challenging objects.
\newblock In \emph{Proceedings of the IEEE/CVF Conference on Computer Vision and Pattern Recognition}, pages 21222--21231, 2022.

\bibitem[Xiang et~al.(2018)Xiang, Schmidt, Narayanan, and Fox]{posecnn}
Yu Xiang, Tanner Schmidt, Venkatraman Narayanan, and Dieter Fox.
\newblock Posecnn: A convolutional neural network for 6d object pose estimation in cluttered scenes.
\newblock \emph{Robotics: Science and Systems (RSS)}, 2018.

\bibitem[Yen-Chen et~al.(2021)Yen-Chen, Florence, Barron, Rodriguez, Isola, and Lin]{iNerf}
Lin Yen-Chen, Pete Florence, Jonathan~T. Barron, Alberto Rodriguez, Phillip Isola, and Tsung-Yi Lin.
\newblock {iNeRF}: Inverting neural radiance fields for pose estimation.
\newblock In \emph{IEEE/RSJ International Conference on Intelligent Robots and Systems ({IROS})}, 2021.

\bibitem[Yen-Chen et~al.(2022)Yen-Chen, Florence, Barron, Lin, Rodriguez, and Isola]{Nerfsupervision}
Lin Yen-Chen, Pete Florence, Jonathan~T. Barron, Tsung-Yi Lin, Alberto Rodriguez, and Phillip Isola.
\newblock {NeRF-Supervision}: Learning dense object descriptors from neural radiance fields.
\newblock In \emph{IEEE Conference on Robotics and Automation ({ICRA})}, 2022.

\bibitem[Zakharov et~al.(2019)Zakharov, Shugurov, and Ilic]{dpod}
Sergey Zakharov, Ivan Shugurov, and Slobodan Ilic.
\newblock Dpod: 6d pose object detector and refiner.
\newblock In \emph{International Conference on Computer Vision (ICCV)}, 2019.

\bibitem[Zhai et~al.(2023)Zhai, Huang, Wu, Jung, Di, Manhardt, Tombari, Navab, and Busam]{zhai2023monograspnet}
Guangyao Zhai, Dianye Huang, Shun-Cheng Wu, Hyunjun Jung, Yan Di, Fabian Manhardt, Federico Tombari, Nassir Navab, and Benjamin Busam.
\newblock Monograspnet: 6-dof grasping with a single rgb image.
\newblock In \emph{IEEE Conference on Robotics and Automation ({ICRA})}, 2023.

\end{thebibliography}
}

\end{document}